\documentclass[10pt,twocolumn,letterpaper]{article}

\usepackage[pagenumbers]{noname} 
\usepackage{algorithm}
\usepackage{algpseudocode}
\usepackage{multirow}
\usepackage{booktabs}
\usepackage{fontawesome5}
\usepackage[table]{xcolor}
\usepackage{colortbl}
\usepackage{graphicx}
\usepackage{lipsum}
\usepackage{pifont}
\usepackage{xparse}
\usepackage{pgf}

\definecolor{heatred}{RGB}{248,105,107}
\definecolor{heatyellow}{RGB}{255,235,132}
\definecolor{heatgreen}{RGB}{99,190,123}

\NewDocumentCommand{\HH}{O{} m m m}{%
  \pgfmathsetmacro{\score}{100*(#2-#3)/(#4-#3)}%
  \pgfmathsetmacro{\score}{min(100,max(0,\score))}%
  \pgfmathtruncatemacro{\s}{round(\score)}%
  \ifnum\s<50
    \pgfmathtruncatemacro{\mix}{100-2*\s}%
    \edef\heatcol{heatred!\mix!heatyellow}%
  \else
    \pgfmathtruncatemacro{\mix}{200-2*\s}%
    \edef\heatcol{heatyellow!\mix!heatgreen}%
  \fi
  \edef\heatcolsoft{\heatcol!45!white}%
  \expandafter\cellcolor\expandafter{\heatcolsoft}%
  \if\relax\detokenize{#1}\relax #2\else #1\fi
}

\NewDocumentCommand{\HL}{O{} m m m}{%
  \pgfmathsetmacro{\score}{100*(#4-#2)/(#4-#3)}%
  \pgfmathsetmacro{\score}{min(100,max(0,\score))}%
  \pgfmathtruncatemacro{\s}{round(\score)}%
  \ifnum\s<50
    \pgfmathtruncatemacro{\mix}{100-2*\s}%
    \edef\heatcol{heatred!\mix!heatyellow}%
  \else
    \pgfmathtruncatemacro{\mix}{200-2*\s}%
    \edef\heatcol{heatyellow!\mix!heatgreen}%
  \fi
  \edef\heatcolsoft{\heatcol!45!white}%
  \expandafter\cellcolor\expandafter{\heatcolsoft}%
  \if\relax\detokenize{#1}\relax #2\else #1\fi
}

\NewDocumentCommand{\HM}{O{} m m m m}{%
  \pgfmathsetmacro{\val}{#2*#5}%
  \pgfmathsetmacro{\mn}{#3*#5}%
  \pgfmathsetmacro{\mx}{#4*#5}%
  \pgfmathprintnumber[fixed,precision=2]{\val}%
}

\NewDocumentCommand{\HHM}{O{} m m m m}{%
  \pgfmathsetmacro{\val}{#2*#5}%
  \pgfmathsetmacro{\mn}{#3*#5}%
  \pgfmathsetmacro{\mx}{#4*#5}%
  \pgfmathsetmacro{\score}{100*(\val-\mn)/(\mx-\mn)}%
  \pgfmathsetmacro{\score}{min(100,max(0,\score))}%
  \pgfmathtruncatemacro{\s}{round(\score)}%
  \ifnum\s<50
    \pgfmathtruncatemacro{\mix}{100-2*\s}%
    \edef\heatcol{heatred!\mix!heatyellow}%
  \else
    \pgfmathtruncatemacro{\mix}{200-2*\s}%
    \edef\heatcol{heatyellow!\mix!heatgreen}%
  \fi
  \edef\heatcolsoft{\heatcol!45!white}%
  \expandafter\cellcolor\expandafter{\heatcolsoft}%
  \if\relax\detokenize{#1}\relax
    \pgfmathprintnumber[fixed,fixed zerofill,precision=2]{\val}%
  \else
    #1%
  \fi
}

\NewDocumentCommand{\HLM}{O{} m m m m}{%
  \pgfmathsetmacro{\val}{#2*#5}%
  \pgfmathsetmacro{\mn}{#3*#5}%
  \pgfmathsetmacro{\mx}{#4*#5}%
  \pgfmathsetmacro{\score}{100*(\mx-\val)/(\mx-\mn)}%
  \pgfmathsetmacro{\score}{min(100,max(0,\score))}%
  \pgfmathtruncatemacro{\s}{round(\score)}%
  \ifnum\s<50
    \pgfmathtruncatemacro{\mix}{100-2*\s}%
    \edef\heatcol{heatred!\mix!heatyellow}%
  \else
    \pgfmathtruncatemacro{\mix}{200-2*\s}%
    \edef\heatcol{heatyellow!\mix!heatgreen}%
  \fi
  \edef\heatcolsoft{\heatcol!45!white}%
  \expandafter\cellcolor\expandafter{\heatcolsoft}%
  \if\relax\detokenize{#1}\relax
    \pgfmathprintnumber[fixed,fixed zerofill,precision=2]{\val}%
  \else
    #1%
  \fi
}

\definecolor{heatred}{RGB}{248,105,107}
\definecolor{heatyellow}{RGB}{255,235,132}
\definecolor{heatgreen}{RGB}{99,190,123}

\newcommand{\RankCellFour}[3][]{%
  \pgfmathtruncatemacro{\mysign}{#2 == 0 ? 0 : (#2 > 0 ? 1 : -1)}%
  \ifnum\mysign=0\relax
     \cellcolor{white}#1
  \else\ifnum\mysign=1\relax
     \pgfmathsetmacro{\ratio}{abs(#2)/abs(#3)}%
     \pgfmathtruncatemacro{\pct}{min(100, max(0, \ratio * 100))}%
     \pgfmathtruncatemacro{\mix}{45 + \pct * 0.55}%
     \edef\heatcol{heatgreen!\mix!white}%
     \edef\heatcolsoft{\heatcol!45!white}%
     \expandafter\cellcolor\expandafter{\heatcolsoft}#1%
  \else
     \pgfmathsetmacro{\ratio}{abs(#2)/abs(#3)}%
     \pgfmathtruncatemacro{\pct}{min(100, max(0, \ratio * 100))}%
     \ifnum\pct<50
        \pgfmathtruncatemacro{\mix}{45 + \pct * 1.10}%
        \edef\heatcol{heatyellow!\mix!white}%
     \else
        \pgfmathtruncatemacro{\mix}{(\pct - 50) * 2}%
        \edef\heatcol{heatred!\mix!heatyellow}%
     \fi
     \edef\heatcolsoft{\heatcol!45!white}%
     \expandafter\cellcolor\expandafter{\heatcolsoft}#1%
  \fi\fi
}

\newcommand{\RankInline}[3][]{%
  \pgfmathtruncatemacro{\mysign}{#2 == 0 ? 0 : (#2 > 0 ? 1 : -1)}%
  {\fboxsep=2pt\relax%
   \ifnum\mysign=0\relax
     #1%
   \else\ifnum\mysign=1\relax
     \pgfmathsetmacro{\ratio}{abs(#2)/abs(#3)}%
     \pgfmathtruncatemacro{\pct}{min(100, max(0, \ratio * 100))}%
     \pgfmathtruncatemacro{\mix}{45 + \pct * 0.55}%
     \edef\heatcol{heatgreen!\mix!white}%
     \edef\heatcolsoft{\heatcol!45!white}%
     \expandafter\colorbox\expandafter{\heatcolsoft}{#1}%
   \else
     \pgfmathsetmacro{\ratio}{abs(#2)/abs(#3)}%
     \pgfmathtruncatemacro{\pct}{min(100, max(0, \ratio * 100))}%
     \ifnum\pct<50
        \pgfmathtruncatemacro{\mix}{45 + \pct * 1.10}%
        \edef\heatcol{heatyellow!\mix!white}%
     \else
        \pgfmathtruncatemacro{\mix}{(\pct - 50) * 2}%
        \edef\heatcol{heatred!\mix!heatyellow}%
     \fi
     \edef\heatcolsoft{\heatcol!45!white}%
     \expandafter\colorbox\expandafter{\heatcolsoft}{#1}%
   \fi\fi}%
}

\definecolor{myblue}{rgb}{0.21,0.49,0.74}
\usepackage[pagebackref,breaklinks,colorlinks,allcolors=myblue]{hyperref}

\title{Resonant Brane Splatting for Arbitrary-Scale Super-Resolution}

\author{
Giulio Federico$^{1,2}$\thanks{Corresponding author.} \quad
Giuseppe Amato$^{2}$ \quad
Claudio Gennaro$^{2}$ \quad
Fabio Carrara$^{2}$ \quad
Marco Di Benedetto$^{2}$ \\
\\
$^1$University of Pisa, Pisa, Italy \quad
$^2$ISTI-CNR, Pisa, Italy \\
{\tt\small \{giulio.federico, giuseppe.amato, claudio.gennaro, fabio.carrara, marco.dibenedetto\}@isti.cnr.it} \\
{\tt\small Code is available at: \url{https://github.com/GiulioFede/ResonantBraneSplatting.git}}
}

\begin{document}
\maketitle
\begin{abstract}
Arbitrary-Scale Super-Resolution (ASR) reconstructs images at continuous magnification factors. Recent methods accelerate inference by replacing computationally heavy implicit neural decoders with explicit 2D Gaussian Splatting (GS). However, since standard Gaussians are smooth low-pass primitives, modeling edges and fine textures requires multiple overlapping, well-aligned splats, which creates severe bottlenecks during rasterization.
%
To address this, we introduce \emph{Resonant Brane Splatting} (RBS), a feed-forward ASR framework. RBS replaces flat Gaussians with \emph{Branes}: expressive primitives that emit spatially varying colors to natively model local contrast and complex textures within a single footprint. We achieve this by augmenting the standard Gaussian envelope with internal Gaussian-Hermite modes, assigning a distinct color coefficient to each. The zero-order mode recovers standard GS, while higher-order modes capture high frequencies.
We predict Brane parameters directly from low-resolution features. Because Branes provide a mathematically richer formulation than simple Gaussians, far fewer primitives need to overlap to reconstruct a given target pixel. To exploit this, we introduce an efficient \emph{fully differentiable rasterizer} with a precise culling strategy based on the classical quantum turning point. This allows us to safely skip negligible regions, drastically reducing the rendering overhead. Experiments on standard ASR benchmarks show that RBS improves reconstruction quality over implicit and GS baselines,
while achieving superior speed-quality trade-off than prior GS methods.
\end{abstract}

\section{Introduction}
\label{sec:intro}

\begin{figure}[t]
\centering
\includegraphics[width=\columnwidth]{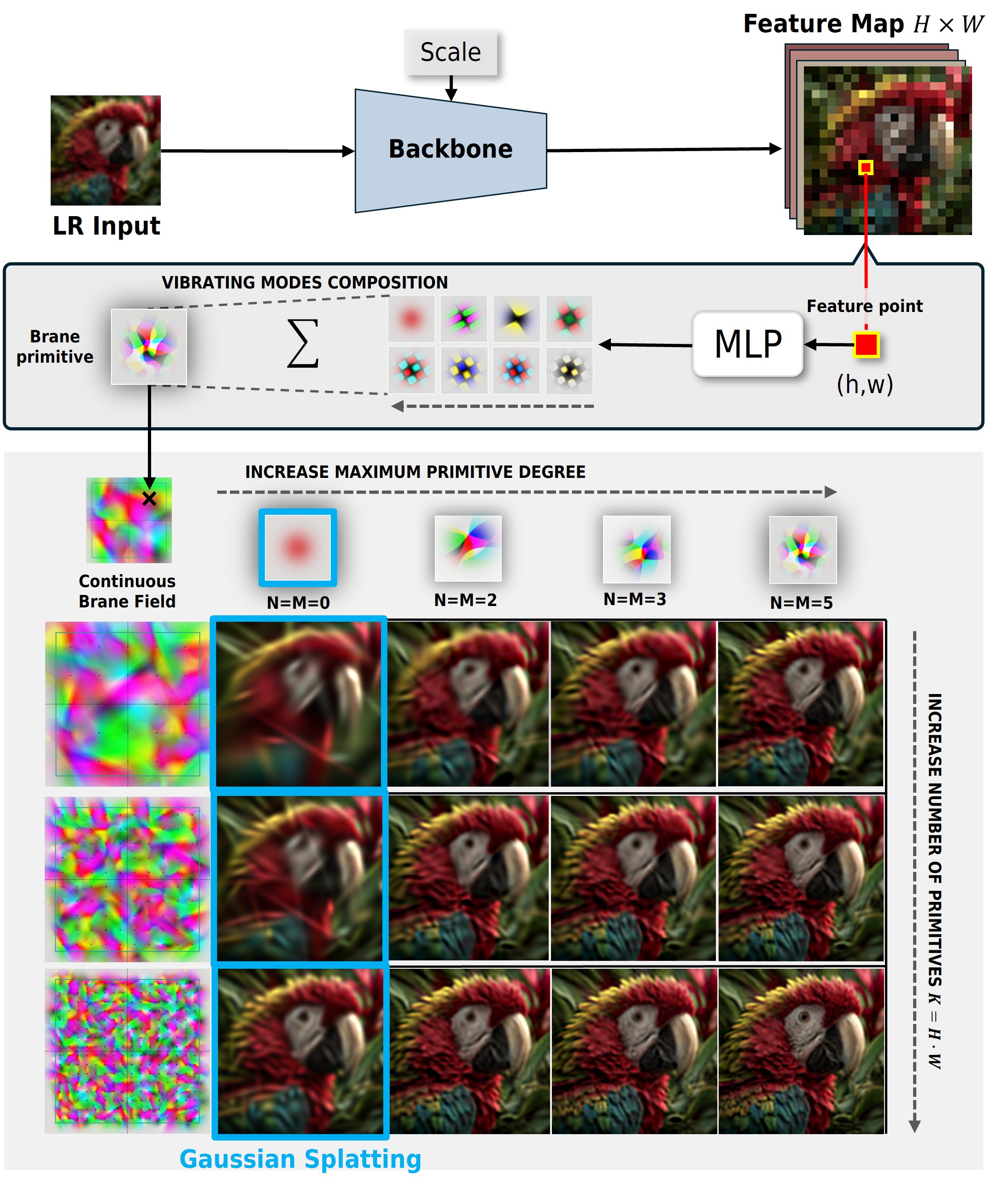}
\caption{\textbf{Resonant Brane Splatting Overview.} Given an LR input and scale factor, a backbone extracts a feature map decoded into our proposed Brane primitives, composing a continuous Brane field. Increasing primitive count and Brane complexity better models SR outputs. Conversely, degree-0 Branes (Gaussian Splatting) yield blurry results under equal primitive budgets.}
\label{fig:rbs_and_mstring_methaphor}
\end{figure}


Image Super-Resolution (SR) recovers details from low-resolution (LR) inputs, while Arbitrary-Scale SR (ASR) extends this goal to continuous magnification factors. Implicit Neural Representations (INRs)~\citep{hu2019meta,chen2021learning,lee2022local,Wei_2023_CVPR,Yao_2023_CVPR,cao2023ciaosr,He_2024_CVPR} dominated ASR by learning coordinate-to-color mappings. While flexible, INRs require dense pixel-wise queries at inference time and suffer from spectral bias, making high frequencies difficult to reconstruct faithfully.


Recent 2D Gaussian Splatting (GS) methods~\citep{hu2025gaussiansr, Chen_2025_ICCV} replace implicit decoding with explicit rendering, predicting primitives from LR features. While efficient, a standard Gaussian remains a smooth, low-pass footprint limited to a single color. Because a single Gaussian cannot represent internal color variations, modeling sharp edges or complex textures forces the network to rely on the precise alignment and collaboration of numerous overlapping splats. This high degree of overlap creates a severe bottleneck, directly slowing down the rasterization process.

To name our proposed 2D splatting framework for ASR (Fig. \ref{fig:rbs_and_mstring_methaphor}), we draw a purely conceptual analogy from M-Theory \cite{m_theory}, where vibrating multidimensional \emph{p-branes} \cite{brane_theory} generalize strings \cite{polchinski1995dirichlet}. 
Our \emph{Brane} primitive adopts the exact mathematical eigenfunctions of the Quantum Harmonic Oscillator \cite{quantum_reference}, augmenting a Gaussian envelope with internal Gaussian-Hermite modes, that ``vibrate'' to create local structural patterns. Ultimately, the aggregation of these patterns constructs the complex visual matter of the super-resolved image.

While prior work applied Gaussian-Hermite functions to spatial support~\cite{2dgh} on 3D tasks, we uniquely use this basis for appearance generation. By assigning distinct color coefficients to each mode, a single Brane natively emits spatially varying colors. The zero-order mode recovers standard Gaussian splatting, while higher-order modes model signed color residuals. A single Brane can therefore synthesize local contrast, sharp edges, and multi-colored textures, representing high-frequency patterns with fewer splats.
A feed-forward network predicts Brane parameters directly from LR features. Since Branes internally model complex textures, dense primitive overlapping is no longer required. To maximize inference speed, we design a custom differentiable CUDA rasterizer that bounds primitive evaluation using the \emph{quantum turning point}, the exact radial threshold before Gaussian-Hermite modes exponentially decay. Safely culling Branes outside this boundary accelerates rendering well beyond standard GS.

Our contributions are: 
(i) \emph{Resonant Brane Splatting}, a Gaussian-Hermite primitive generalizing flat, single-color Gaussians into structured, color-generating splats; 
(ii) a feed-forward ASR framework predicting Brane geometry, opacity and mode-wise color coefficients from LR features; 
(iii) a differentiable 2D GPU/CUDA-based scale-aware rasterizer exploiting bounded Gaussian-Hermite support for faster inference; and 
(iv) state-of-the-art visual performance, establishing a new optimal balance between rendering speed and image quality.

\begin{figure*}
\centering
\includegraphics[width=\textwidth]{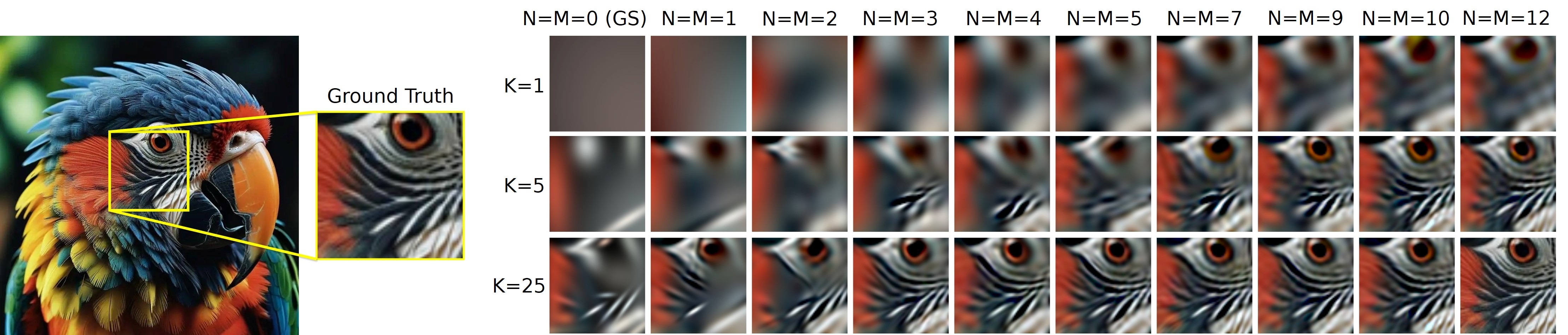}
\caption{\textbf{Brane Expressiveness.} Left: Cropped ground truth. Right: Reconstructions varying the number of Branes ($K$) and Gaussian-Hermite degrees ($N, M$), where $N=M=0$ corresponds to standard Gaussian Splatting. Please zoom in for details.}

\label{fig:checkboard_example}
\end{figure*}

\section{Related Work}

\paragraph{Image Super-Resolution.}
Fixed-scale image Super-Resolution has evolved from early CNNs~\cite{dong2014learning} to sophisticated architectures leveraging expanded receptive fields~\citep{Kim_2016_CVPR, kim2016deeply, lai2017deep}, compressed feature spaces~\cite{dong2016accelerating}, and Residual-in-Residual structures~\citep{wang2018esrgan, zhang2018image, zhang2018residual, Li_2019_CVPR}. 
Transformer models~\citep{Liang_2021_ICCV, zhang2022efficient, zhang2024transcending} further improved long-range dependency modeling. While diffusion models excel in perceptual SR, iterative sampling and hallucination risks hinder efficient, faithful reconstruction ~\cite{chen2024ssl, liang2022details, wu2024seesr, yi2025fine, yu2024scaling}. Crucially, these methods require scale-specific training and lack continuous magnification at inference.

\paragraph{Arbitrary-Scale SR and Continuous Representations.}
ASR breaks integer-scale constraints by framing SR as a continuous representation problem~\cite{hu2019meta, wang2021learning}. LIIF~\cite{chen2021learning} formalized this by mapping continuous spatial coordinates to RGB values via an MLP. Subsequent works mitigated spectral bias and improved feature aggregation through Fourier mappings~\cite{lee2022local}, Normalizing Flows~\cite{Yao_2023_CVPR}, implicit attention~\cite{cao2023ciaosr}, neural operators~\cite{Wei_2023_CVPR}, and frequency encodings~\cite{Chen_2023_CVPR, NEURIPS2021_6e7d5d25}. Because dense point-wise MLP queries create severe inference bottlenecks, LMF~\cite{He_2024_CVPR} confined heavy computation to the LR space. However, coordinate-wise decoding still treats each output pixel independently, limiting data reuse across neighboring pixels and motivating explicit rendering alternatives based on spatial primitives.


\paragraph{Splatting-Based ASR.}
Recently, 2D Gaussian Splatting (GS)~\cite{kerbl20233d} emerged as a faster ASR alternative. GaussianSR~\cite{hu2025gaussiansr} blends pre-trained kernels, while GSASR~\cite{Chen_2025_ICCV} and the lightweight GRAPE~\cite{grape} predict learnable anisotropic parameters. ContinuousSR~\cite{continuoussr} explicitly models continuous HR signals in a single pass. However, standard Gaussians inherently act as low-pass footprints requiring strong anisotropy for high-frequency synthesis, limiting local expressiveness and making reconstruction sensitive to splat alignment.


\paragraph{Expressive Splatting Primitives.}
To overcome GS limitations, recent works explore modified primitives \citep{beta_splatting, deblur_3dgs, ges, 3dls, 3dcs, drk}. Recognizing that the standard flat-color assumption severely restricts textural fidelity, methods like Gaussian Billboards \cite{gaussianbillboards} and GStex \cite{GStex} replace the single color with discrete texture grids. Alternatively, 2DGH \cite{2dgh}, closely related to our work, introduces Gaussian-Hermite kernels to modulate the spatial footprint itself, increasing the kernel's capacity to represent sharp boundaries and fine geometric structures during scene optimization.

\paragraph{Positioning of our work.}
We build on the expressive Gaussian-Hermite primitive but differ in how appearance is represented. In 2DGH~\cite{2dgh}, the Gaussian-Hermite function defines the spatial opacity/support profile, while color remains a separate splat attribute. In contrast, we assign an RGB-valued coefficient to each Hermite mode, so the color emitted by a Brane varies according to its internal modes rather than being constant over the primitive. This allows a single primitive to model local contrast, structure, and multi-colored texture within the same footprint, instead of relying on stretched shapes or multiple overlapping splats to approximate high-frequency details.
\section{Method}
\label{sec:method}

\subsection{Limitations of Splatting-Based ASR}

Given a low-resolution input image $\mathbf{I}_{LR} \in \mathbb{R}^{H \times W \times 3}$ and a continuous target scale factor $s \in \mathbb{R}^{+}$, Arbitrary-Scale Super-Resolution predicts an output image $\mathbf{I}_{SR} \in \mathbb{R}^{H_\text{SR} \times W_\text{SR} \times 3}$ with $H_\text{SR} = \lfloor Hs \rfloor$ and $W_\text{SR} = \lfloor Ws \rfloor$. We follow the splatting-based ASR framework of GSASR~\cite{Chen_2025_ICCV}: 
a neural backbone extracts a dense feature map $\hat{\mathbf{F}}\in\mathbb{R}^{\hat H \times \hat W \times C}$. To increase its spatial density, it can be optionally upsampled by a factor $u\in\mathbb{N}^{+}$, obtaining $\mathbf{F}\in\mathbb{R}^{H' \times W' \times C}$, with
$H'=u\hat H$ and $W'=u\hat W$. Each feature vector in $\mathbf{F}$ predicts one spatial primitive to be rendered at the target resolution.
Let $K=H'W'$ be the number of primitives and let $\mathbf{f}_k \in \mathbb{R}^{C}$ denote the feature associated with the $k$-th grid cell. Each primitive is anchored in the normalized image domain $[0,1]^2$ and contributes to the final color $\mathbf{C}(u,v)$ of a target pixel $(u,v)$ through explicit rasterization.
%
In standard Gaussian splatting,
\begin{equation}
    \mathbf{C}(u,v) = \sum_{k=1}^{K} \alpha_k E(x'_k,y'_k)\mathbf{c}_k\,,
    \label{eq:gs_primitive}
\end{equation}
where each primitive $k$ has a single color vector $\mathbf{c}_k \in [0,1]^3$, an opacity $\alpha_k$, a Gaussian spatial envelope $E(x,y)=\exp\left(-\frac{1}{2}(x^2+y^2)\right)$ computed at the local canonical coordinates $(x'_k,y'_k)$ of pixel $(u,v)$. 

This formulation is efficient, but the primitive is locally limited: the color is constant over the footprint and all high-frequency structure must be explained through geometry and overlap with neighboring splats.

\subsection{Resonant Brane Primitive}

To overcome this bottleneck, we replace each flat Gaussian splat with a \textbf{Resonant Brane}. A Brane keeps the Gaussian envelope of Eq.~\ref{eq:gs_primitive}, but augments it with internal Gaussian-Hermite modes. Just as these modes describe the localized vibrational states of a Quantum Harmonic Oscillator \cite{quantum_reference}, they define higher-order polynomial patterns inside the exact same spatial support. This allows the primitive to represent complex internal structure instead of behaving as a flat, single-color footprint.

\begin{figure}[t]
\centering
\includegraphics[width=\columnwidth]{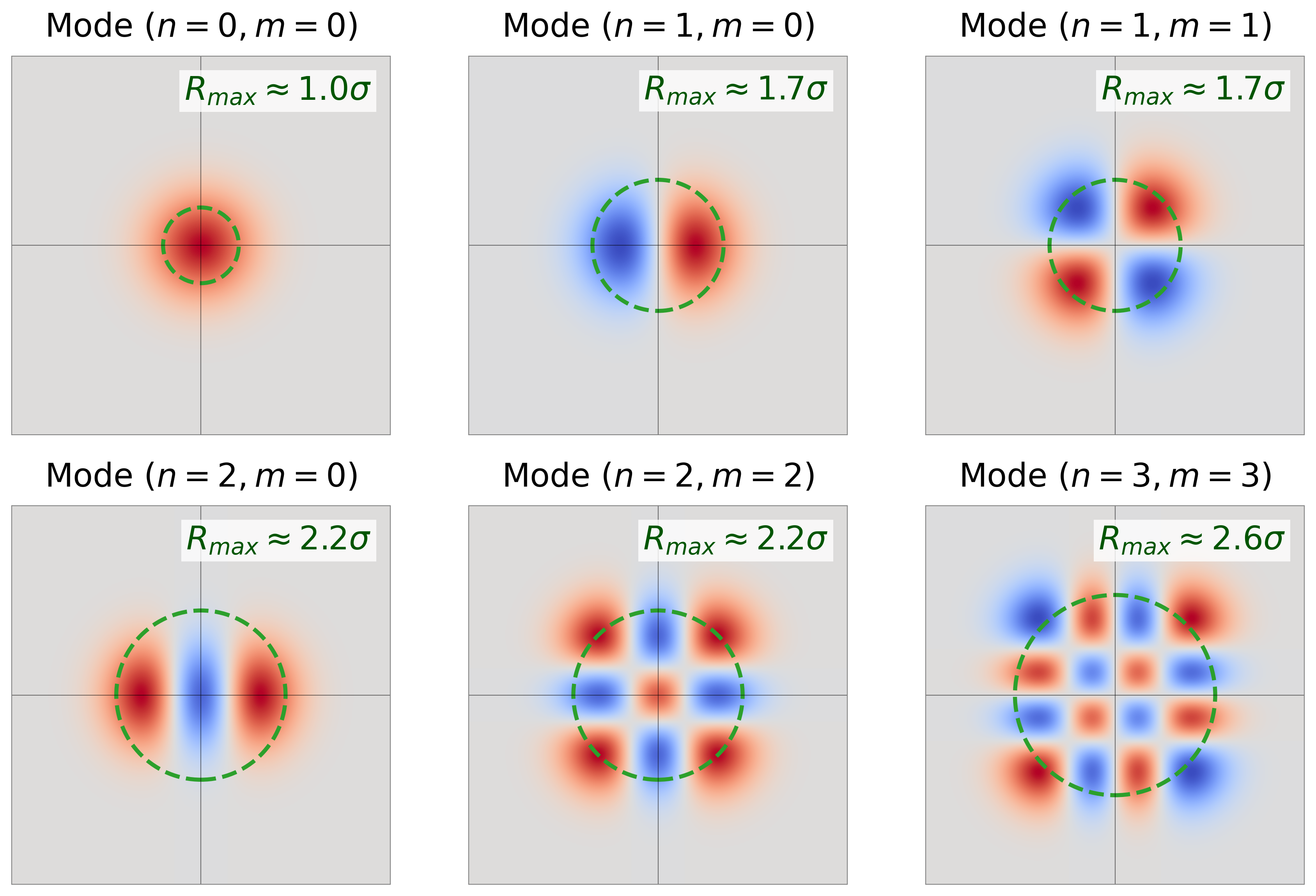}
\caption{\textbf{Visualization of 2D Gaussian-Hermite modes}. The zero-order mode recovers the standard Gaussian splat, while higher-order modes introduce signed spatial variations within the same footprint. For each mode, the turning point $R_\text{max}$ is shown in dashed green; beyond this radius, the mode contribution decays exponentially.}
\label{fig:vibrant_modes}
\end{figure}

We define the Brane in a local 2D canonical space. The spatial boundary is governed by the same Gaussian envelope $E(x,y)$. Inside this envelope, we use normalized Hermite polynomials \cite{hermite_polynomials}
\begin{equation}
\begin{aligned}
\bar{H}_n(z) &= \frac{1}{\sqrt{2^n n!}} H_n(z), \quad \text{where}\\
H_0(z) &= 1, \qquad H_1(z)=2z, \quad \text{and} \\
H_n(z) &= 2zH_{n-1}(z)-2(n-1)H_{n-2}(z)\,\text{ for }\,n>1\,. \\
\end{aligned}
\label{eq:normalized_hermite_recurrence}
\end{equation}
Here, $H_n(z)$ is the physicist's Hermite polynomial of degree $n$, that can be computed efficiently via the above recursive relation \cite{hermite_expansions}. The normalization removes the dominant degree-dependent growth of $H_n$, improving numerical stability when using higher-order modes.

Local 2D variations inside the Gaussian envelope can be expressed by combining one-dimensional Hermite modes along the canonical axes. A 2D mode of degree $(n,m)$ is given by the tensor product
\begin{equation}
    \Psi_{n,m}(x,y)=\bar{H}_n(x)\bar{H}_m(y)\,.
    \label{eq:hermite_mode}
\end{equation}
Figure~\ref{fig:vibrant_modes} shows 2D modes when varying the degree $(n,m)$.

Gaussian-Hermite functions have already been used in splatting to make the spatial support (i.e., the shape of the opacity kernel) more expressive than an ellipse~\cite{2dgh}. We use the same modal basis but to \emph{model appearance}. Instead of assigning one color to the whole primitive, we assign its own color coefficient $\mathbf{c}_{n,m} \in \mathbb{R}^3$ to each mode. The Brane emission is therefore
\begin{equation}
    \mathcal{B}(x,y)
    =
    \alpha E(x,y)
    \sum_{n=0}^{N}
    \sum_{m=0}^{M}
    \mathbf{c}_{n,m}\Psi_{n,m}(x,y) \,.
    \label{eq:brane_definition}
\end{equation}
This makes a Brane a color-generating primitive. The zero-order mode ($N=M=0$) gives $\mathcal{B}(x,y) = \alpha E(x,y) \mathbf{c}_{0,0}$, 
which is exactly a standard Gaussian splat. Higher-order modes instead act as signed color residuals. They can add or subtract color locally inside the same footprint, allowing one primitive to synthesize local contrast, edges, and multi-colored textures without decomposing them into many small or highly anisotropic Gaussians (Fig. \ref{fig:checkboard_example}).

\subsection{Predicting Branes from LR Features}

We now map the Brane primitive to ASR. The feature grid $\mathbf{F}$ defines $K=H'W'$ Branes over the normalized image domain. For each feature vector $\mathbf{f}_k$, the network predicts the full Brane state
\begin{equation}
\Phi_k =
\left[
\begin{array}{c@{\;}c@{\;}c}
\underbrace{\Delta x_k,\Delta y_k}_{\text{anchoring}}, &
\underbrace{\sigma_{x,k},\sigma_{y,k},\rho_k,\alpha_k}_{\text{footprint \& opacity}}, &
\underbrace{\mathbf{c}_{0,0,k},\ldots,\mathbf{c}_{N,M,k}}_{\text{mode colors}}
\end{array}
\right].
\label{eq:brane_parameters}
\end{equation}

\noindent\textbf{Geometric anchoring.}
The $k$-th Brane is initially anchored at the center of its feature grid cell
\begin{equation}
    (x^{\text{ref}}_k,y^{\text{ref}}_k)
    =
    \left(
    \frac{j+0.5}{W'},
    \frac{i+0.5}{H'}
    \right).
\end{equation}
To allow sub-pixel adaptation, an MLP head predicts continuous offsets
\begin{equation}
    (\Delta x_k,\Delta y_k)=\operatorname{MLP}_{\theta_1}(\mathbf{f}_{k})\,,
\end{equation}
yielding the final center $(x_k, y_k)$ as
\begin{equation}
    x_k=x^{\text{ref}}_k+\frac{\Delta x_k}{W'}\,,
    \qquad
    y_k=y^{\text{ref}}_k+\frac{\Delta y_k}{H'}\,.
\end{equation}

\begin{figure}[t]
\centering
\includegraphics[width=\columnwidth]{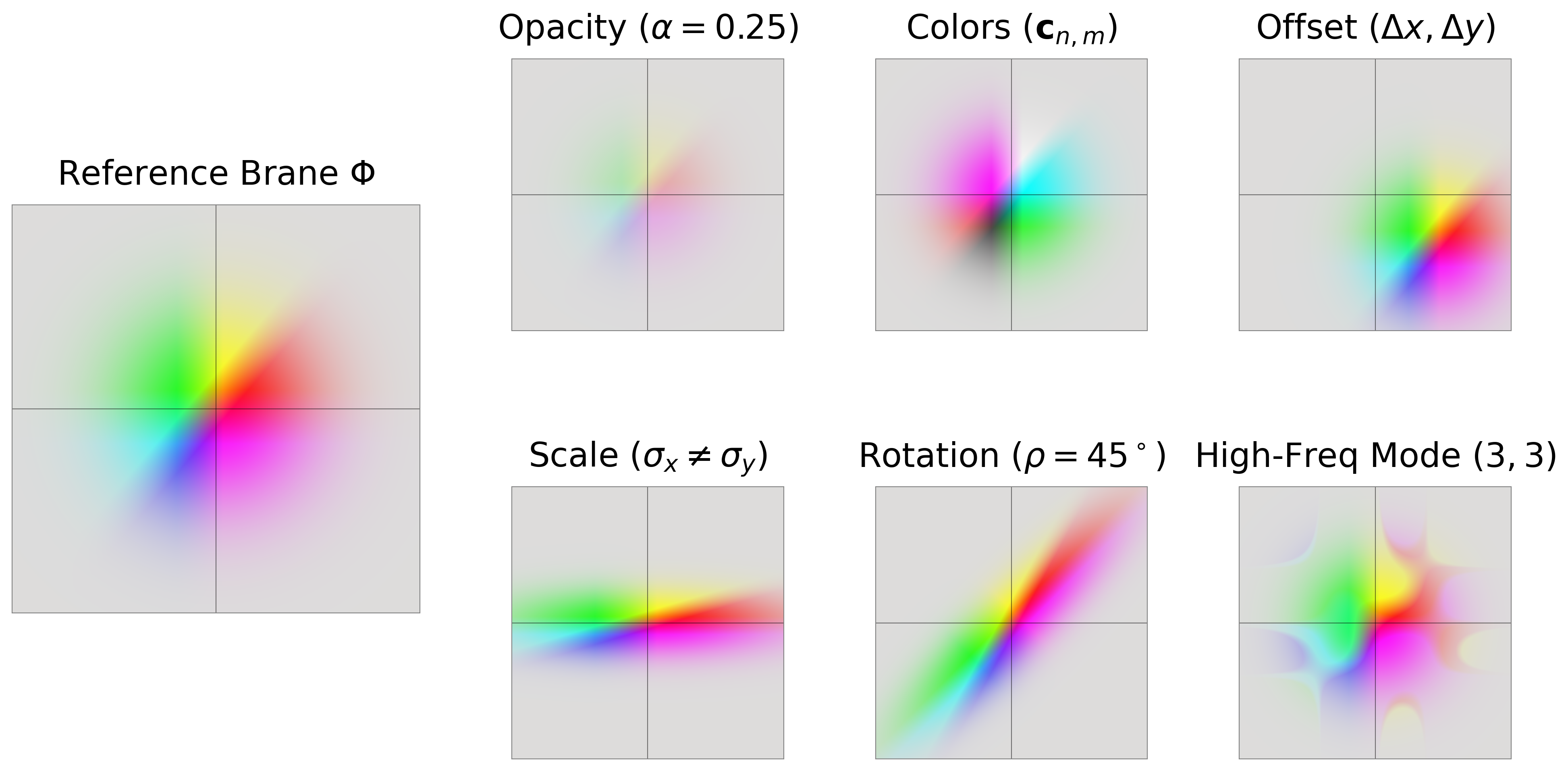}
\caption{\textbf{Effect of Brane parameters.} Mode colors control the internal appearance of the primitive, while footprint and opacity control where and how strongly the Brane contributes.}
\label{fig:brane_parameters}
\end{figure}

\noindent\textbf{Footprint and opacity.}
Additional heads predict the anisotropic scales, rotation, and opacity as
\begin{align}
    (\sigma_{x,k},\sigma_{y,k})
    &=
    \operatorname{sigmoid}(\operatorname{MLP}_{\theta_3}(\mathbf{f}_{k})),\\
    \rho_k
    &=
    \pi\,\operatorname{tanh}(\operatorname{MLP}_{\theta_4}(\mathbf{f}_{k})),\\
    \alpha_k
    &=
    \operatorname{sigmoid}(\operatorname{MLP}_{\theta_5}(\mathbf{f}_{k})).
\end{align}
Figure~\ref{fig:brane_parameters} shows the effect of parameters on appearance.

\noindent\textbf{Mode colors.}
Given maximum degrees $N$ and $M$, each Brane has $L=(N+1)(M+1)$ modes. A dedicated head predicts all raw mode colors
\begin{equation}
    \tilde{\mathbf{C}}_k
    =
    \operatorname{MLP}_{\theta_2}(\mathbf{f}_{k})
    \in \mathbb{R}^{3L}.
\end{equation}
We reshape this vector into $L$ color coefficients $\tilde{\mathbf{c}}_{k}^{n,m}\in\mathbb{R}^3$. The zero-order coefficient is constrained to a positive color range, while higher-order modes are signed residuals:
\begin{equation}
    \mathbf{c}_{k}^{n,m} =
    \begin{cases}
        \operatorname{sigmoid}(\tilde{\mathbf{c}}_{k}^{n,m}) ,
        & n=0,\ m=0,\\
        \operatorname{tanh}(\tilde{\mathbf{c}}_{k}^{n,m}) ,
        & \text{otherwise}.
    \end{cases}
    \label{eq:mode_color_activation}
\end{equation}
Thus, $\mathbf{c}^{0,0}_{k}$ defines the smooth base color, while higher-order coefficients can add or subtract color inside the footprint.

\subsection{Differentiable Rasterization}

\begin{figure*}[t]
\centering
\includegraphics[width=0.9\textwidth,trim={0 0.05cm 0 0},clip]{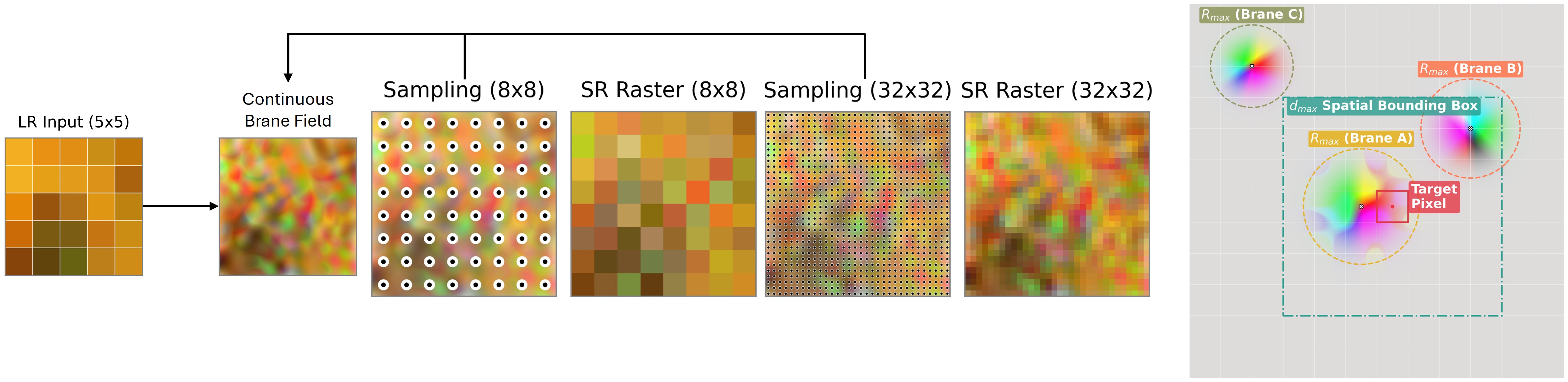}

\caption{\textbf{Rasterization and culling.} Decoded LR features form a continuous Brane space where sampling grid resolution dictates SR resolution (left). Target pixel color is computed by culling primitives outside box $d_{\max}$ or beyond quantum turning point $R_{\max}$ (right).}
\label{fig:rasterization_process}
\end{figure*}

To render the output image, each target pixel $(u,v)$ is first projected into normalized coordinates
\begin{equation}
    (u_{\text{norm}},v_{\text{norm}})
    =
    \left(
    \frac{u+0.5}{W_\text{SR}},
    \frac{v+0.5}{H_\text{SR}}
    \right).
\end{equation}
For the $k$-th Brane, we compute the displacement from its center as
\begin{equation}
    (\delta x_k,\delta y_k)
    =
    (u_{\text{norm}}-x_k,\ v_{\text{norm}}-y_k)
\end{equation}
that is then mapped to the local canonical coordinates through inverse rotation and scaling
\begin{equation}
    \begin{pmatrix}
    x'_k\\
    y'_k
    \end{pmatrix}
    =
    \begin{pmatrix}
    \frac{1}{\sigma_{x,k}} & 0\\
    0 & \frac{1}{\sigma_{y,k}}
    \end{pmatrix}
    \begin{pmatrix}
    \cos\rho_k & \sin\rho_k\\
    -\sin\rho_k & \cos\rho_k
    \end{pmatrix}
    \begin{pmatrix}
    \delta x_k\\
    \delta y_k
    \end{pmatrix}.
    \label{eq:inverse_transformation}
\end{equation}
The final RGB value $\mathbf{C}(u,v)$ is obtained by additively splatting all active Branes
\begin{equation}
    \mathbf{C}(u,v)
    =
    \sum_{k=1}^{K}
    \alpha_k
    E(x'_k,y'_k)
    \sum_{n=0}^{N}
    \sum_{m=0}^{M}
    \mathbf{c}_{n,m,k}
    \bar{H}_n(x'_k)
    \bar{H}_m(y'_k).
    \label{eq:rbs_rendering}
\end{equation}


\paragraph{Culling Branes.} Dense evaluation across all Branes and pixels requires $\mathcal{O}(K H_{SR} W_{SR})$ operations. Our CUDA rasterizer (Algorithm~\ref{alg:scale_aware_rasterization}) avoids this via a dual culling strategy to efficiently determine the final color of each pixel. First, we map the target pixel into the continuous Brane field and define a bounding region around it using $d_{\max}$, which represents a specific fraction of this normalized space. We restrict our evaluation strictly to the candidate Branes located inside this area. Second, to accelerate rendering further, we filter these surviving Branes by evaluating only those with a meaningful contribution. We achieve this using a canonical space boundary check based on the quantum turning point $R_{\max} = \sqrt{2n_{\max} + 1}$ 
\cite{quantum_reference}, that is the radius beyond which the visual impact of a Brane with maximum degree $n_{\max}=\max{(N,M)}$ is empirically negligible. 
Thus, we discard candidates where ${x'_k}^2 + {y'_k}^2 > R_{\max}^2$.


\begin{algorithm}
\caption{Differentiable CUDA rasterization of Resonant Branes}
\label{alg:scale_aware_rasterization}
\begin{algorithmic}[1]
\Require Branes $\{\mathcal{B}_k\}_{k=1}^{K}$, output size $(H_{SR},W_{SR})$, bounding box $d_{\max}$, maximum degree $n_{\max}$
\Ensure Rendered image $\mathbf{I}_{SR}$
\State Initialize $\mathbf{I}_{SR}$ as zeros
\State $R_{\max}^2 \gets 2n_{\max}+1$
\For{each Brane $\mathcal{B}_k$}

        \State Initialize $\Phi_k$ from $\mathcal{B}_k$

    \For{each pixel $(u,v)$ in $\mathbf{I}_{SR}$}
        \State $u_{\text{norm}},v_{\text{norm}} \gets \frac{u+0.5}{sW}, \frac{v+0.5}{sH}$
        \State $\delta x_k, \delta y_k  \gets u_{\text{norm}} - x_k, v_{\text{norm}} - y_k$
        
        \If{$|\delta x_k| < d_{\max}$ \textbf{and} $|\delta y_k| < d_{\max}$}

             \State Obtain $x_k', y'_k$ using Eq.~\ref{eq:inverse_transformation} 
             
             \If{${x'_k}^2 + {y'_k}^2 \le R_{\max}^2$} 
                 \State Obtain $\mathcal{B}_k(x_k', y'_k)$ using Eq.~\ref{eq:brane_definition}
                 \State $I_{SR}(u,v) \gets I_{SR}(u,v) + \mathcal{B}_k(x_k', y'_k)$
             \EndIf
        \EndIf
    \EndFor
\EndFor

\end{algorithmic}
\end{algorithm}

\section{Experiments}
\label{sec:esperiments}

\newcommand{\Pa}[2][]{\if\relax\detokenize{#1}\relax\HH{#2}{30.08}{33.91}\else\HH[#1]{#2}{30.08}{33.91}\fi}
\newcommand{\Pb}[2][]{\if\relax\detokenize{#1}\relax\HH{#2}{27.61}{29.93}\else\HH[#1]{#2}{27.61}{29.93}\fi}
\newcommand{\Pc}[2][]{\if\relax\detokenize{#1}\relax\HH{#2}{25.86}{27.62}\else\HH[#1]{#2}{25.86}{27.62}\fi}
\newcommand{\Pd}[2][]{\if\relax\detokenize{#1}\relax\HH{#2}{23.51}{24.86}\else\HH[#1]{#2}{23.51}{24.86}\fi}
\newcommand{\Pe}[2][]{\if\relax\detokenize{#1}\relax\HH{#2}{22.24}{23.33}\else\HH[#1]{#2}{22.24}{23.33}\fi}
\newcommand{\Pf}[2][]{\if\relax\detokenize{#1}\relax\HH{#2}{20.74}{21.51}\else\HH[#1]{#2}{20.74}{21.51}\fi}

\newcommand{\Sa}[2][]{\if\relax\detokenize{#1}\relax\HHM{#2}{0.9169}{0.9429}{100}\else\HHM[#1]{#2}{0.9169}{0.9429}{100}\fi}
\newcommand{\Sb}[2][]{\if\relax\detokenize{#1}\relax\HHM{#2}{0.8365}{0.8833}{100}\else\HHM[#1]{#2}{0.8365}{0.8833}{100}\fi}
\newcommand{\Sc}[2][]{\if\relax\detokenize{#1}\relax\HHM{#2}{0.7806}{0.8274}{100}\else\HHM[#1]{#2}{0.7806}{0.8274}{100}\fi}
\newcommand{\Sd}[2][]{\if\relax\detokenize{#1}\relax\HHM{#2}{0.6659}{0.7305}{100}\else\HHM[#1]{#2}{0.6659}{0.7305}{100}\fi}
\newcommand{\Se}[2][]{\if\relax\detokenize{#1}\relax\HHM{#2}{0.5970}{0.6606}{100}\else\HHM[#1]{#2}{0.5970}{0.6606}{100}\fi}
\newcommand{\Sf}[2][]{\if\relax\detokenize{#1}\relax\HHM{#2}{0.5185}{0.5746}{100}\else\HHM[#1]{#2}{0.5185}{0.5746}{100}\fi}

\newcommand{\La}[2][]{\if\relax\detokenize{#1}\relax\HLM{#2}{0.0470}{0.0667}{100}\else\HLM[#1]{#2}{0.0470}{0.0667}{100}\fi}
\newcommand{\Lb}[2][]{\if\relax\detokenize{#1}\relax\HLM{#2}{0.1202}{0.1541}{100}\else\HLM[#1]{#2}{0.1202}{0.1541}{100}\fi}
\newcommand{\Lc}[2][]{\if\relax\detokenize{#1}\relax\HLM{#2}{0.1838}{0.2322}{100}\else\HLM[#1]{#2}{0.1838}{0.2322}{100}\fi}
\newcommand{\Ld}[2][]{\if\relax\detokenize{#1}\relax\HLM{#2}{0.2790}{0.3283}{100}\else\HLM[#1]{#2}{0.2790}{0.3283}{100}\fi}
\newcommand{\Le}[2][]{\if\relax\detokenize{#1}\relax\HLM{#2}{0.3570}{0.4327}{100}\else\HLM[#1]{#2}{0.3570}{0.4327}{100}\fi}
\newcommand{\Lf}[2][]{\if\relax\detokenize{#1}\relax\HLM{#2}{0.4788}{0.5929}{100}\else\HLM[#1]{#2}{0.4788}{0.5929}{100}\fi}

\newcommand{\Da}[2][]{\if\relax\detokenize{#1}\relax\HLM{#2}{0.0595}{0.0759}{100}\else\HLM[#1]{#2}{0.0595}{0.0759}{100}\fi}
\newcommand{\Db}[2][]{\if\relax\detokenize{#1}\relax\HLM{#2}{0.1069}{0.1376}{100}\else\HLM[#1]{#2}{0.1069}{0.1376}{100}\fi}
\newcommand{\Dc}[2][]{\if\relax\detokenize{#1}\relax\HLM{#2}{0.1438}{0.1721}{100}\else\HLM[#1]{#2}{0.1438}{0.1721}{100}\fi}
\newcommand{\Dd}[2][]{\if\relax\detokenize{#1}\relax\HLM{#2}{0.2053}{0.2340}{100}\else\HLM[#1]{#2}{0.2053}{0.2340}{100}\fi}
\newcommand{\De}[2][]{\if\relax\detokenize{#1}\relax\HLM{#2}{0.2460}{0.2759}{100}\else\HLM[#1]{#2}{0.2460}{0.2759}{100}\fi}
\newcommand{\Df}[2][]{\if\relax\detokenize{#1}\relax\HLM{#2}{0.3072}{0.3485}{100}\else\HLM[#1]{#2}{0.3072}{0.3485}{100}\fi}

\begin{table*}[t]
\centering
\resizebox{\textwidth}{!}{
\begin{tabular}{l cccccc cccccc cccccc cccccc}
\toprule
\multirow{2}{*}{Method} & \multicolumn{6}{c}{PSNR $\uparrow$} & \multicolumn{6}{c}{SSIM $\uparrow$ ($\times 100$)} & \multicolumn{6}{c}{LPIPS $\downarrow$ ($\times 100$)} & \multicolumn{6}{c}{DISTS $\downarrow$ ($\times 100$)} \\
\cmidrule(lr){2-7} \cmidrule(lr){8-13} \cmidrule(lr){14-19} \cmidrule(lr){20-25}
 & $\times 2$ & $\times 3$ & $\times 4$ & $\times 6$ & $\times 8$ & $\times 12$ 
 & $\times 2$ & $\times 3$ & $\times 4$ & $\times 6$ & $\times 8$ & $\times 12$ 
 & $\times 2$ & $\times 3$ & $\times 4$ & $\times 6$ & $\times 8$ & $\times 12$ 
 & $\times 2$ & $\times 3$ & $\times 4$ & $\times 6$ & $\times 8$ & $\times 12$ \\
\midrule
\multicolumn{25}{c}{\textit{Implicit Methods}} \\
\midrule
Meta-SR  & \Pa{33.04} & \Pb{28.94} & \Pc{26.71} & \Pd{24.07} & \Pe{22.64} & \Pf{21.00} & \Sa{0.9363} & \Sb{0.8677} & \Sc{0.8055} & \Sd{0.6966} & \Se{0.6213} & \Sf{0.5317} & \La{0.0552} & \Lb{0.1381} & \Lc{0.2062} & \Ld{0.3158} & \Le{0.4022} & \Lf{0.5332} & \Da{0.0666} & \Db{0.1194} & \Dc{0.1562} & \Dd{0.2084} & \De{0.2512} & \Df{0.3144} \\
LIIF  & \Pa{32.84} & \Pb{28.81} & \Pc{26.67} & \Pd{24.19} & \Pe{22.78} & \Pf{21.15} & \Sa{0.9353} & \Sb{0.8664} & \Sc{0.8041} & \Sd{0.7028} & \Se{0.6338} & \Sf{0.5499} & \La{0.0569} & \Lb{0.1382} & \Lc{0.2077} & \Ld{0.3099} & \Le{0.3903} & \Lf{0.5198} & \Da{0.0666} & \Db{0.1194} & \Dc{0.1612} & \Dd{0.2176} & \De{0.2591} & \Df{0.3209} \\
LTE  & \Pa{33.00} & \Pb{28.96} & \Pc{26.80} & \Pd{24.27} & \Pe{22.86} & \Pf{21.22} & \Sa{0.9365} & \Sb{0.8686} & \Sc{0.8074} & \Sd{0.7058} & \Se{0.6362} & \Sf{0.5514} & \La{0.0552} & \Lb{0.1373} & \Lc{0.2047} & \Ld{0.3220} & \Le{0.4107} & \Lf{0.5477} & \Da{0.0660} & \Db{0.1186} & \Dc{0.1600} & \Dd{0.2160} & \De{0.2586} & \Df{0.3216} \\
SRNO  & \Pa{33.27} & \Pb{29.12} & \Pc{26.97} & \Pd{24.42} & \Pe{23.01} & \Pf{21.35} & \Sa{0.9390} & \Sb{0.8714} & \Sc{0.8119} & \Sd{0.7113} & \Se{0.6423} & \Sf{0.5571} & \La{0.0518} & \Lb{0.1332} & \Lc{0.1987} & \Ld{0.3148} & \Le{0.4011} & \Lf{0.5374} & \Da{0.0640} & \Db{0.1158} & \Dc{0.1563} & \Dd{0.2110} & \De{0.2520} & \Df{0.3149} \\
LINF  & \Pa{32.87} & \Pb{28.82} & \Pc{26.69} & \Pd{24.18} & \Pe{22.77} & \Pf{21.12} & \Sa{0.9350} & \Sb{0.8658} & \Sc{0.8039} & \Sd{0.7010} & \Se{0.6305} & \Sf{0.5462} & \La{0.0569} & \Lb{0.1399} & \Lc{0.2090} & \Ld{0.3155} & \Le{0.4010} & \Lf{0.5365} & \Da{0.0671} & \Db{0.1208} & \Dc{0.1636} & \Dd{0.2217} & \De{0.2657} & \Df{0.3304} \\
LMF  & \Pa{33.08} & \Pb{29.11} & \Pc{26.94} & \Pd{24.39} & \Pe{22.97} & \Pf{21.33} & \Sa{0.9371} & \Sb{0.8709} & \Sc{0.8104} & \Sd{0.7102} & \Se{0.6408} & \Sf{0.5561} & \La{0.0557} & \Lb{0.1335} & \Lc{0.2020} & \Ld{0.3169} & \Le{0.4037} & \Lf{0.5403} & \Da{0.0659} & \Db{0.1167} & \Dc{0.1589} & \Dd{0.2149} & \De{0.2580} & \Df{0.3216} \\
Ciao-SR  & \Pa{33.30} & \Pb{29.17} & \Pc{27.10} & \Pd{24.58} & \Pe{23.13} & \Pf{21.44} & \Sa{0.9388} & \Sb{0.8716} & \Sc{0.8142} & \Sd{0.7173} & \Se{0.6484} & \Sf{0.5637} & \La{0.0534} & \Lb{0.1334} & \Lc{0.1966} & \Ld{0.2950} & \Le{0.3733} & \Lf{0.5005} & \Da{0.0638} & \Db{0.1134} & \Dc{0.1559} & \Dd{0.2111} & \De{0.2507} & \Df{0.3114} \\
\midrule
\multicolumn{25}{c}{\textit{Explicit Methods}} \\
\midrule
Gaussian-SR  & \Pa{32.96} & \Pb{28.93} & \Pc{26.77} & \Pd{24.16} & \Pe{22.64} & \Pf{20.84} & \Sa{0.9363} & \Sb{0.8680} & \Sc{0.8064} & \Sd{0.6996} & \Se{0.6234} & \Sf{0.5319} & \La{0.0563} & \Lb{0.1384} & \Lc{0.2069} & \Ld{0.3241} & \Le{0.4249} & \Lf{0.5929} & \Da{0.0665} & \Db{0.1196} & \Dc{0.1610} & \Dd{0.2191} & \De{0.2704} & \Df{0.3485} \\
GSASR  & \Pa{33.53} & \Pb{29.35} & \Pc{27.15} & \Pd{24.63} & \Pe{23.19} & \Pf{21.47} & \Sa{0.9406} & \Sb{0.8760} & \Sc{0.8177} & \Sd{0.7214} & \Se{0.6522} & \Sf{0.5670} & \La{0.0507} & \Lb{0.1294} & \Lc{0.1953} & \Ld{0.2943} & \Le{0.3723} & \Lf{0.4927} & \Da{0.0617} & \Db{0.1126} & \Dc{0.1515} & \Dd{0.2064} & \De[\textbf{24.60}]{0.2460} & \Df[\textbf{30.72}]{0.3072} \\
GRAPE  & \Pa{30.08} & \Pb{27.61} & \Pc{25.86} & \Pd{23.51} & \Pe{22.24} & \Pf{20.74} & \Sa{0.9169} & \Sb{0.8365} & \Sc{0.7806} & \Sd{0.6659} & \Se{0.5970} & \Sf{0.5185} & \La{0.0667} & \Lb{0.1541} & \Lc{0.2322} & \Ld{0.3283} & \Le{0.4327} & \Lf{0.5848} & \Da{0.0759} & \Db{0.1376} & \Dc{0.1721} & \Dd{0.2340} & \De{0.2759} & \Df{0.3380} \\
CSR*  &  -  & \Pb{28.48} & \Pc{26.50} & \Pd{24.25} & \Pe{22.95} & \Pf{21.41} &  -  & \Sb{0.8683} & \Sc{0.8104} & \Sd{0.7183} & \Se{0.6505} & \Sf{0.5662} &  -  & \Lb{0.1356} & \Lc{0.2080} & \Ld{0.3037} & \Le{0.3982} & \Lf{0.5236} &  -  & \Db{0.1149} & \Dc{0.1541} & \Dd{0.2098} & \De{0.2487} & \Df{0.3084} \\

\midrule
\textbf{RBS ($\mathbf{u}=1$) (ours)}  & \Pa{33.73} & \Pb{29.67} & \Pc{27.30} & \Pd{24.56} & \Pe{23.10} & \Pf{21.38} & \Sa{0.9420} & \Sb{0.8796} & \Sc{0.8195} & \Sd{0.7173} & \Se{0.6462} & \Sf{0.5607} & \La{0.0489} & \Lb{0.1224} & \Lc{0.1942} & \Ld{0.3179} & \Le{0.4084} & \Lf{0.5502} & \Da{0.0605} & \Db[\textbf{10.69}]{0.1069} & \Dc[\textbf{14.38}]{0.1438} & \Dd[\textbf{20.53}]{0.2053} & \De{0.2521} & \Df{0.3191} \\
\textbf{RBS ($\mathbf{u}=2$)  (ours)}  & \Pa[\textbf{33.91}]{33.91} & \Pb[\textbf{29.93}]{29.93} & \Pc[\textbf{27.62}]{27.62} & \Pd[\textbf{24.86}]{24.86} & \Pe{23.32} & \Pf[\textbf{21.51}]{21.51} & \Sa[\textbf{94.29}]{0.9429} & \Sb[\textbf{88.33}]{0.8833} & \Sc[\textbf{82.74}]{0.8274} & \Sd{0.7296} & \Se{0.6588} & \Sf{0.5719} & \La{0.0483} & \Lb{0.1222} & \Lc{0.1873} & \Ld{0.2863} & \Le{0.3653} & \Lf{0.4895} & \Da{0.0611} & \Db{0.1112} & \Dc{0.1507} & \Dd{0.2078} & \De{0.2487} & \Df{0.3099} \\
\textbf{RBS ($\mathbf{u}=4$)  (ours)}  & \Pa{33.81} & \Pb{29.85} & \Pc{27.58} & \Pd[\textbf{24.86}]{24.86} & \Pe[\textbf{23.33}]{23.33} & \Pf[\textbf{21.51}]{21.51} & \Sa{0.9428} & \Sb{0.8829} & \Sc{0.8272} & \Sd[\textbf{73.05}]{0.7305} & \Se[\textbf{66.06}]{0.6606} & \Sf[\textbf{57.46}]{0.5746} & \La[\textbf{4.70}]{0.0470} & \Lb[\textbf{12.02}]{0.1202} & \Lc[\textbf{18.38}]{0.1838} & \Ld[\textbf{27.90}]{0.2790} & \Le[\textbf{35.70}]{0.3570} & \Lf[\textbf{47.88}]{0.4788} & \Da[\textbf{5.95}]{0.0595} & \Db{0.1099} & \Dc{0.1505} & \Dd{0.2099} & \De{0.2514} & \Df{0.3112} \\
\bottomrule
\end{tabular}
}
\caption{Quantitative results of Implicit and Explicit ASR models on Urban100 \cite{urban100}. All models use the RDN \cite{zhang2018residual} backbone, except GRAPE \cite{grape} (EDSR \cite{Lim_2017_CVPR_Workshops}) and ContinuousSR \cite{continuoussr} (CSR*) (HAT \cite{hat_backbone}),  as the only provided by their authors. We omit the $\times 2$ results for CSR* (-) due to a bug in its architecture at this specific scale. Best results are in \textbf{bold}; PSNR and SSIM are computed on the Y channel.}
\label{tab:quantitative_results_urban100_rdn}
\end{table*}

\begin{table}[htbp]
\centering

\newcommand{\ITa}[2][]{\if\relax\detokenize{#1}\relax\HL{#2}{0.07}{100.0}\else\HL[#1]{#2}{0.07}{100.0}\fi}
\newcommand{\ITb}[2][]{\if\relax\detokenize{#1}\relax\HL{#2}{0.48}{100.0}\else\HL[#1]{#2}{0.48}{100.0}\fi}
\newcommand{\ITc}[2][]{\if\relax\detokenize{#1}\relax\HL{#2}{0.67}{100.0}\else\HL[#1]{#2}{0.67}{100.0}\fi}
\newcommand{\ITd}[2][]{\if\relax\detokenize{#1}\relax\HL{#2}{0.8}{100.0}\else\HL[#1]{#2}{0.8}{100.0}\fi}
\newcommand{\ITe}[2][]{\if\relax\detokenize{#1}\relax\HL{#2}{0.85}{100.0}\else\HL[#1]{#2}{0.85}{100.0}\fi}
\newcommand{\ITf}[2][]{\if\relax\detokenize{#1}\relax\HL{#2}{0.71}{100.0}\else\HL[#1]{#2}{0.71}{100.0}\fi}

\newcommand{\GMa}[2][]{\if\relax\detokenize{#1}\relax\HL{#2}{2.53}{100.0}\else\HL[#1]{#2}{2.53}{100.0}\fi}
\newcommand{\GMb}[2][]{\if\relax\detokenize{#1}\relax\HL{#2}{6.05}{100.0}\else\HL[#1]{#2}{6.05}{100.0}\fi}
\newcommand{\GMc}[2][]{\if\relax\detokenize{#1}\relax\HL{#2}{4.66}{100.0}\else\HL[#1]{#2}{4.66}{100.0}\fi}
\newcommand{\GMd}[2][]{\if\relax\detokenize{#1}\relax\HL{#2}{3.7}{100.0}\else\HL[#1]{#2}{3.7}{100.0}\fi}
\newcommand{\GMe}[2][]{\if\relax\detokenize{#1}\relax\HL{#2}{2.2}{100.0}\else\HL[#1]{#2}{2.2}{100.0}\fi}
\newcommand{\GMf}[2][]{\if\relax\detokenize{#1}\relax\HL{#2}{1.1}{100.0}\else\HL[#1]{#2}{1.1}{100.0}\fi}

\setlength{\tabcolsep}{3.5pt}
\resizebox{\linewidth}{!}{%
\begin{tabular}{l rrrrrr rrrrrr}
\toprule
\multirow{2}{*}{Methods} 
& \multicolumn{6}{c}{Inference Time (ms) $\downarrow$} 
& \multicolumn{6}{c}{GPU Memory (MB) $\downarrow$} \\
\cmidrule(lr){2-7} \cmidrule(lr){8-13}
& \multicolumn{1}{c}{$\times 2$} 
& \multicolumn{1}{c}{$\times 3$} 
& \multicolumn{1}{c}{$\times 4$} 
& \multicolumn{1}{c}{$\times 6$} 
& \multicolumn{1}{c}{$\times 8$} 
& \multicolumn{1}{c}{$\times 12$}
& \multicolumn{1}{c}{$\times 2$} 
& \multicolumn{1}{c}{$\times 3$} 
& \multicolumn{1}{c}{$\times 4$} 
& \multicolumn{1}{c}{$\times 6$} 
& \multicolumn{1}{c}{$\times 8$} 
& \multicolumn{1}{c}{$\times 12$} \\

\midrule
\multicolumn{13}{c}{\textit{Implicit Methods}} \\
\midrule
Meta-SR & \ITa[178]{0.75} & \ITb[104]{5.04} & \ITc[82]{6.82} & \ITd[69]{9.19} & \ITe[55]{7.83} & \ITf[58]{8.27} & \GMa[8,775]{17.82} & \GMb[8,454]{83.86} & \GMc[8,344]{100.0} & \GMd[8,265]{100.0} & \GMe[8,236]{100.0} & \GMf[8,216]{100.0} \\
LIIF & \ITa[518]{2.18} & \ITb[486]{23.55} & \ITc[220]{18.3} & \ITd[201]{26.76} & \ITe[187]{26.64} & \ITf[178]{25.39} & \GMa[\textbf{1,246}]{2.53} & \GMb[\textbf{610}]{6.05} & \GMc[\textbf{389}]{4.66} & \GMd[345]{4.17} & \GMe[328]{3.98} & \GMf[317]{3.86} \\
LTE & \ITa[254]{1.07} & \ITb[181]{8.77} & \ITc[153]{12.73} & \ITd[136]{18.11} & \ITe[128]{18.23} & \ITf[129]{18.4} & \GMa[1,970]{4.0} & \GMb[930]{9.23} & \GMc[566]{6.78} & \GMd[315]{3.81} & \GMe[302]{3.67} & \GMf[293]{3.57} \\
SRNO & \ITa[236]{1.0} & \ITb[163]{7.9} & \ITc[150]{12.48} & \ITd[118]{15.71} & \ITe[107]{15.24} & \ITf[114]{16.26} & \GMa[6,381]{12.96} & \GMb[6,361]{63.1} & \GMc[6,354]{76.15} & \GMd[6,350]{76.83} & \GMe[6,348]{77.08} & \GMf[6,347]{77.25} \\
LINF & \ITa[202]{0.85} & \ITb[128]{6.2} & \ITc[101]{8.4} & \ITd[84]{11.19} & \ITe[77]{10.97} & \ITf[78]{11.13} & \GMa[3,732]{7.58} & \GMb[3,571]{35.42} & \GMc[3,516]{42.14} & \GMd[3,475]{42.04} & \GMe[3,462]{42.03} & \GMf[3,451]{42.0} \\
LMF & \ITa[311]{1.31} & \ITb[172]{8.33} & \ITc[113]{9.4} & \ITd[94]{12.52} & \ITe[80]{11.4} & \ITf[110]{15.69} & \GMa[7,096]{14.41} & \GMb[6,108]{60.59} & \GMc[5,764]{69.08} & \GMd[5,517]{66.75} & \GMe[5,431]{65.94} & \GMf[5,369]{65.35} \\
CiaoSR & \ITa[23,711]{100.0} & \ITb[2,064]{100.0} & \ITc[1,202]{100.0} & \ITd[736]{98.0} & \ITe[633]{90.17} & \ITf[546]{77.89} & \GMa[49,231]{100.0} & \GMb[10,081]{100.0} & \GMc[3,390]{40.63} & \GMd[1,625]{19.66} & \GMe[1,583]{19.22} & \GMf[1,549]{18.85} \\
\midrule
\multicolumn{13}{c}{\textit{Explicit Methods}} \\
\midrule
GaussianSR & \ITa[1,280]{5.4} & \ITb[955]{46.27} & \ITc[824]{68.55} & \ITd[751]{100.0} & \ITe[702]{100.0} & \ITf[701]{100.0} & \GMa[5,278]{10.72} & \GMb[5,216]{51.74} & \GMc[5,130]{61.48} & \GMd[5,301]{64.14} & \GMe[5,092]{61.83} & \GMf[5,291]{64.4} \\
GSASR & \ITa[1,679]{7.08} & \ITb[857]{41.52} & \ITc[572]{47.59} & \ITd[284]{37.82} & \ITe[206]{29.34} & \ITf[98]{13.98} & \GMa[13,535]{27.49} & \GMb[6,080]{60.31} & \GMc[3,500]{41.95} & \GMd[1,658]{20.06} & \GMe[1,132]{13.74} & \GMf[553]{6.73} \\
CSR* & \ITa[938]{3.96} & \ITb[421]{20.4} & \ITc[273]{22.71} & \ITd[133]{17.71} & \ITe[88]{12.54} & \ITf[69]{9.84} & \GMa[6,042]{12.27} & \GMb[2,868]{28.45} & \GMc[1,918]{22.99} & \GMd[1,070]{12.95} & \GMe[771]{9.36} & \GMf[559]{6.8} \\
GRAPE & \ITa[\textbf{16}]{0.07} & \ITb[\textbf{10}]{0.48} & \ITc[\textbf{8}]{0.67} & \ITd[\textbf{6}]{0.8} & \ITe[\textbf{6}]{0.85} & \ITf[\textbf{5}]{0.71} & \GMa[2,615]{5.31} & \GMb[1,173]{11.64} & \GMc[668]{8.01} & \GMd[\textbf{306}]{3.7} & \GMe[\textbf{181}]{2.2} & \GMf[\textbf{90}]{1.1} \\
\midrule
\textbf{RBS ($\mathbf{u}=1$)  (ours)} & \ITa[433]{1.83} & \ITb[223]{10.8} & \ITc[119]{9.9} & \ITd[64]{8.52} & \ITe[57]{8.12} & \ITf[49]{6.99} & \GMa[2,243]{4.56} & \GMb[1,002]{9.94} & \GMc[701]{8.4} & \GMd[417]{5.05} & \GMe[334]{4.06} & \GMf[246]{2.99} \\
\textbf{RBS ($\mathbf{u}=2$)  (ours)} & \ITa[499]{2.1} & \ITb[242]{11.72} & \ITc[140]{11.65} & \ITd[73]{9.72} & \ITe[62]{8.83} & \ITf[50]{7.13} & \GMa[3,910]{7.94} & \GMb[1,845]{18.3} & \GMc[1,121]{13.43} & \GMd[605]{7.32} & \GMe[459]{5.57} & \GMf[296]{3.6} \\
\textbf{RBS ($\mathbf{u}=4$)  (ours)} & \ITa[701]{2.96} & \ITb[337]{16.33} & \ITc[197]{16.39} & \ITd[92]{12.25} & \ITe[72]{10.26} & \ITf[51]{7.28} & \GMa[13,896]{28.23} & \GMb[6,286]{62.35} & \GMc[3,622]{43.41} & \GMd[1,720]{20.81} & \GMe[1,174]{14.25} & \GMf[578]{7.04} \\
\bottomrule
\end{tabular}%
}
\caption{\textbf{Computational costs.} Evaluated on $720^2$ DIV2K~\cite{Timofte_2017_CVPR_Workshops} HR targets with RDN~\cite{zhang2018residual} (larger scales $\Rightarrow$ smaller inputs $\Rightarrow$ lower costs). We report time (ms) and memory (MB). *ContinuousSR uses HAT~\cite{hat_backbone}; GRAPE uses EDSR~\cite{Lim_2017_CVPR_Workshops}. Best in \textbf{bold}.}
\label{tab:comp_cost_rdn}
\end{table}

\subsection{Experimental Setup}


\paragraph{Implementation details.}
\label{sec:impl_det}
We adopt the GSASR~\cite{Chen_2025_ICCV} pipeline, using EDSR~\cite{Lim_2017_CVPR_Workshops} or RDN~\cite{zhang2018residual} as the baseline extractor. Features are refined by a Transformer and upsampled to \(\mathbf{F}\in \mathbb{R}^{uH\times uW \times 180}\), replacing standard 2D Gaussian projection heads with our Brane parameter heads (Eq.~\ref{eq:brane_parameters}) and using our rasterizer. We test $u \in \{1, 2, 4\}$ to vary the number of generated primitives, isolating the Brane's true expressive power from the capacity of a large decoder. We set Hermite degree $N=M=3$. Crucially, base colors use a Sigmoid activation while oscillatory residuals use Tanh; we avoid hard clipping during additive blending, letting the end-to-end $\mathcal{L}_1$ supervision natively bound the final rendering within the $[0,1]$. For training on DIV2K~\cite{Timofte_2017_CVPR_Workshops}, a $48 \times 48$ LR patch at continuous scale $s$ is rendered to $\lfloor 48s \rfloor^2$ and optimized via $\mathcal{L}_1$ loss against the HR crop. We train with Adam for 500k iterations ($s \sim \mathcal{U}(1, 4)$, initial LR $10^{-4}$ halved every 100k steps), on three H100 GPUs.

\paragraph{Compared models.}
We compare against leading \emph{implicit} (Meta-SR~\cite{hu2019meta}, LIIF~\cite{chen2021learning}, LTE~\cite{lee2022local}, SRNO~\cite{Wei_2023_CVPR}, LINF~\cite{Yao_2023_CVPR}, CiaoSR~\cite{cao2023ciaosr}, LMF~\cite{He_2024_CVPR}) and \emph{explicit} (ContinuousSR~\cite{continuoussr}, GRAPE~\cite{grape}, GaussianSR~\cite{hu2025gaussiansr}, GSASR~\cite{Chen_2025_ICCV}) methods. To explore alternative 3D primitives in 2D ASR, we integrated 2DGH~\cite{2dgh} and Gaussian Billboards (GB)~\cite{gaussianbillboards} into our rasterizer and retrained from scratch. These validate our design: GB contrasts our continuous color with discrete grids, while with 2DGH we prove that Hermite bases favor internal color over spatial deformation.

\paragraph{Evaluation protocol.}
We evaluate on 
Set5~\cite{set5}, Set14~\cite{set14}, DIV2K~\cite{Timofte_2017_CVPR_Workshops}, Urban100~\cite{urban100}, BSDS100~\cite{bsds100}, Manga109~\cite{manga109}, General100~\cite{GENERAL100}, LSDIR~\cite{lsdir_dataset} 
at standard ($s \in \{2, 3, 4, 6, 8, 12\}$) and extreme scales ($s \in \{16, 18, 24, 30\}$, see Supp.), reporting Y-channel PSNR/SSIM~\cite{SSIM}, LPIPS~\cite{lpips}, and DISTS~\cite{dists}. Following
~\cite{Chen_2025_ICCV}, efficiency is benchmarked on an H100 with fixed $720^2$ DIV2K HR crops. 
Larger scales process smaller LR inputs, naturally reducing primitives, memory, and runtime.

\subsection{Discussion}
\label{sec:discussion}

\begin{figure*}[t]
\centering
\includegraphics[width=0.917\textwidth]{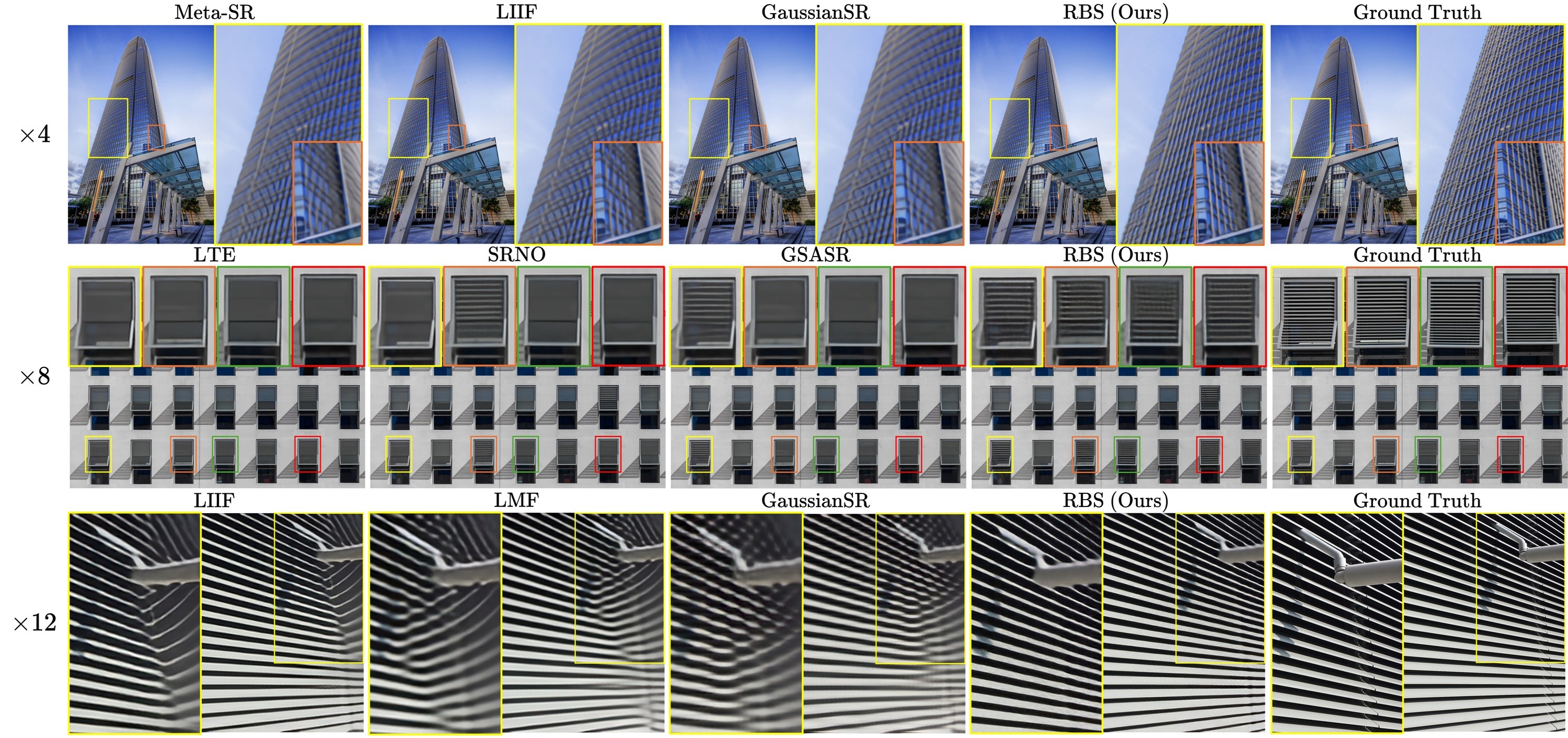}
\caption{\textbf{Qualitative results.} Visual comparison between RBS and several competing methods across different scale factors, using RDN \cite{zhang2018residual} as the backbone network. Detailed results for each model are provided in the Supplementary Material.}

\label{fig:qualitative_results}
\end{figure*}

\paragraph{Brane vs. Gaussian: higher quality at equal primitive count.}
Table~\ref{tab:quantitative_results_urban100_rdn} evaluates RBS for \(u \in \{4, 2, 1\}\) on Urban100~\cite{urban100} --- the hardest benchmark given its intricate structural details (others in Supp.). As explained in Implementation details (Sec. \ref{sec:impl_det}), \(u\) determines the \(C \times uH \times uW\) resolution of the final feature map \(\mathbf{F}\) extracted by the decoder from the LR image. Since each spatial feature of \(\mathbf{F}\) is decoded into a Brane, increasing \(u\) means using more primitives during rasterization. GSASR~\cite{Chen_2025_ICCV} uses \(u=4\); matching it against RBS ($u=4$)  thus ensures a strictly fair comparison under identical decoder capacity and primitive count (\(4H \cdot 4W\)). This isolates our advantage: at equal count, RBS achieves superior metrics with comparable memory (Tab.~\ref{tab:comp_cost_rdn}) and sharper textures (Fig.~\ref{fig:qualitative_results}), proving Branes intrinsically outperform Gaussians.

\begin{figure}[t]
\centering
\includegraphics[width=\columnwidth]{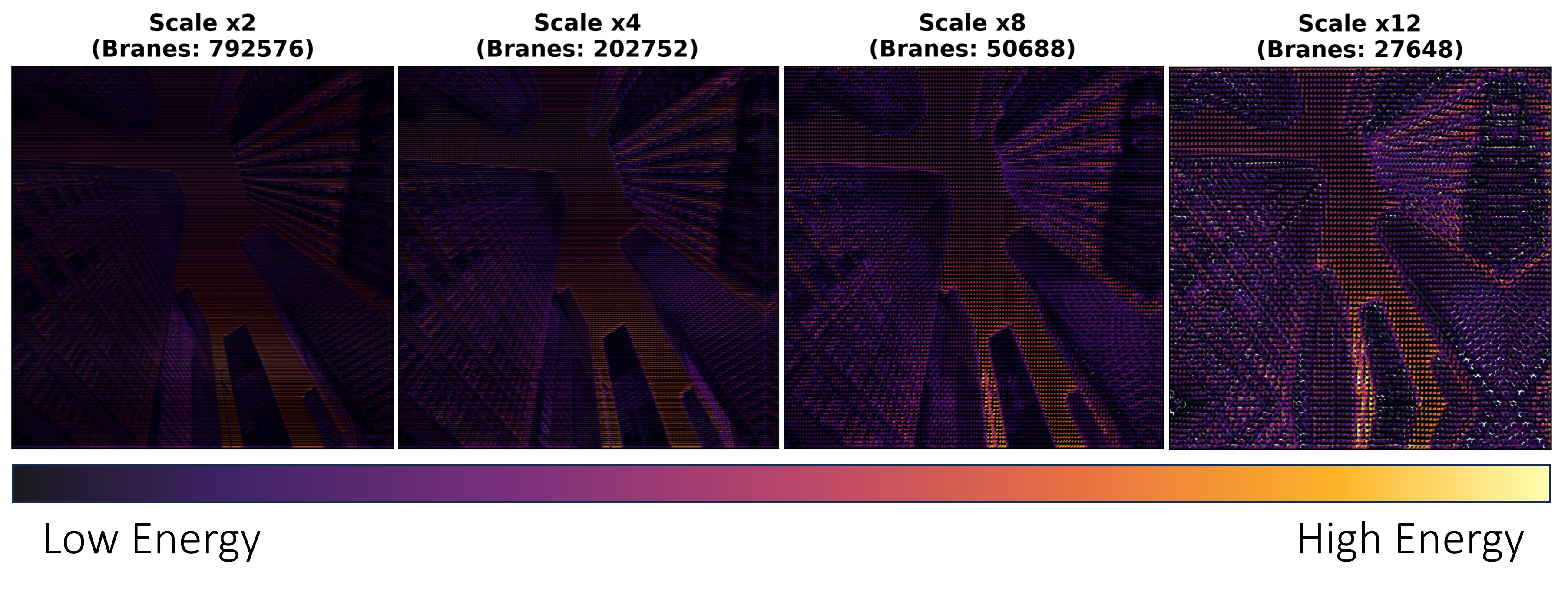}
\caption{\textbf{Brane vibration energy across scales.} Energy is the opacity-weighted ($\alpha$) sum of absolute mode coefficients. As scale increases and primitive counts drop, Branes compensate via higher vibration energy instead of dense overlap, keeping our reduced variants (RBS ($\mathbf{u}=2$)  and RBS ($\mathbf{u}=1$))  highly competitive.}
\label{fig:brane_energy_variation}
\end{figure}

\paragraph{Brane expressiveness enables massive primitive reduction.}
By scaling down the feature upsampling stage from $u=4$ to $u=2$ and $u=1$, we reduce the generated primitives by $4\times$ and $16\times$. To compensate, Branes natively vibrate with higher energy (Fig.~\ref{fig:brane_energy_variation}), capturing complex regions without relying on dense overlap. While standard RBS ($\mathbf{u}=4$)  is required to establish the definitive state-of-the-art across all benchmarks, RBS ($\mathbf{u}=2$)  emerges as the optimal quality-efficiency trade-off (details in Supp. Material). As shown in Table~\ref{tab:comp_cost_rdn}, at the demanding $\times 2$ scale, it drops memory from standard 13.9~GB down to just 3.9~GB. Remarkably, RBS ($\mathbf{u}=1$)  pushes efficiency even further: despite a $16\times$ primitive reduction, it still outperforms GSASR~\cite{Chen_2025_ICCV} at $\times 2$--$\times 4$ scales while slashing the $\times 2$ memory footprint by $\sim 84\%$ (to 2.2~GB), vastly outperforming the lightweight GRAPE~\cite{grape} using similar memory consumption.

\paragraph{Internal expressiveness minimizes overlap and accelerates rendering.}
The speedups in Table~\ref{tab:comp_cost_rdn} stem from alleviating the rasterization bottleneck. Standard Gaussians emit flat colors, requiring heavy overlap to resolve high frequencies. Branes instead model multi-colored textures \textit{internally}, drastically reducing this overlap. Furthermore, because their envelope decays predictably, our CUDA rasterizer strictly discards evaluations beyond the quantum turning point ($R_{\max}$). Applying this exact boundary check to a natively sparser field directly accelerates rendering.

\paragraph{The worst-case scenario: simple benchmarks.}
While RBS excels on complex textures, it struggles on smooth datasets like Set5 (full evaluation in Supp.). Flat Gaussians are natively optimal here, whereas RBS must expend capacity suppressing higher-order modes to collapse Branes into flat envelopes. Consequently (Table~\ref{tab:rbs_vs_gsasr_on_set5}), RBS slightly trails GSASR in PSNR/SSIM, though it surprisingly retains an edge in perceptual metrics (LPIPS/DISTS).




\begin{table}[h!]
\centering

\newcommand{\bestcell}[1]{\cellcolor{green!25}\textbf{#1}}
\newcommand{\worstcell}[1]{\cellcolor{red!25}#1}

\resizebox{\columnwidth}{!}{
\setlength{\tabcolsep}{4pt}
\begin{tabular}{l cccccccc}
\toprule
\multirow{3}{*}{Method} & \multicolumn{2}{c}{$\times 2$} & \multicolumn{2}{c}{$\times 4$} & \multicolumn{2}{c}{$\times 8$} & \multicolumn{2}{c}{$\times 12$} \\
\cmidrule(lr){2-3} \cmidrule(lr){4-5} \cmidrule(lr){6-7} \cmidrule(lr){8-9}
 & PSNR $\uparrow$ & LPIPS $\downarrow$ & PSNR $\uparrow$ & LPIPS $\downarrow$ & PSNR $\uparrow$ & LPIPS $\downarrow$ & PSNR $\uparrow$ & LPIPS $\downarrow$ \\
 & SSIM $\uparrow$ & DISTS $\downarrow$ & SSIM $\uparrow$ & DISTS $\downarrow$ & SSIM $\uparrow$ & DISTS $\downarrow$ & SSIM $\uparrow$ & DISTS $\downarrow$ \\
\midrule
\multirow{2}{*}{GSASR} 
 & \bestcell{38.38} & \worstcell{0.0510} & \bestcell{32.83} & \worstcell{0.1688} & \bestcell{27.52} & \worstcell{0.2788} & \bestcell{24.68} & \worstcell{0.3813} \\
 & \bestcell{0.9621} & \worstcell{0.0767} & \bestcell{0.9027} & \worstcell{0.1551} & \bestcell{0.7937} & \worstcell{0.2285} & \worstcell{0.7053} & \worstcell{0.2820} \\
\midrule
\multirow{2}{*}{RBS ($\mathbf{u}=4$)} 
 & \worstcell{38.28} & \bestcell{0.0508} & \worstcell{32.77} & \bestcell{0.1646} & \worstcell{27.44} & \bestcell{0.2747} & \worstcell{24.65} & \bestcell{0.3755} \\
 & \worstcell{0.9618} & \bestcell{0.0755} & \worstcell{0.9021} & \bestcell{0.1535} & \worstcell{0.7924} & \bestcell{0.2262} & \bestcell{0.7082} & \bestcell{0.2814} \\
\bottomrule
\end{tabular}
}
\caption{\textbf{Worst-case scenario on Set5.}}
\label{tab:rbs_vs_gsasr_on_set5}
\end{table}

\begin{figure}[t]
\centering

\includegraphics[width=\columnwidth]{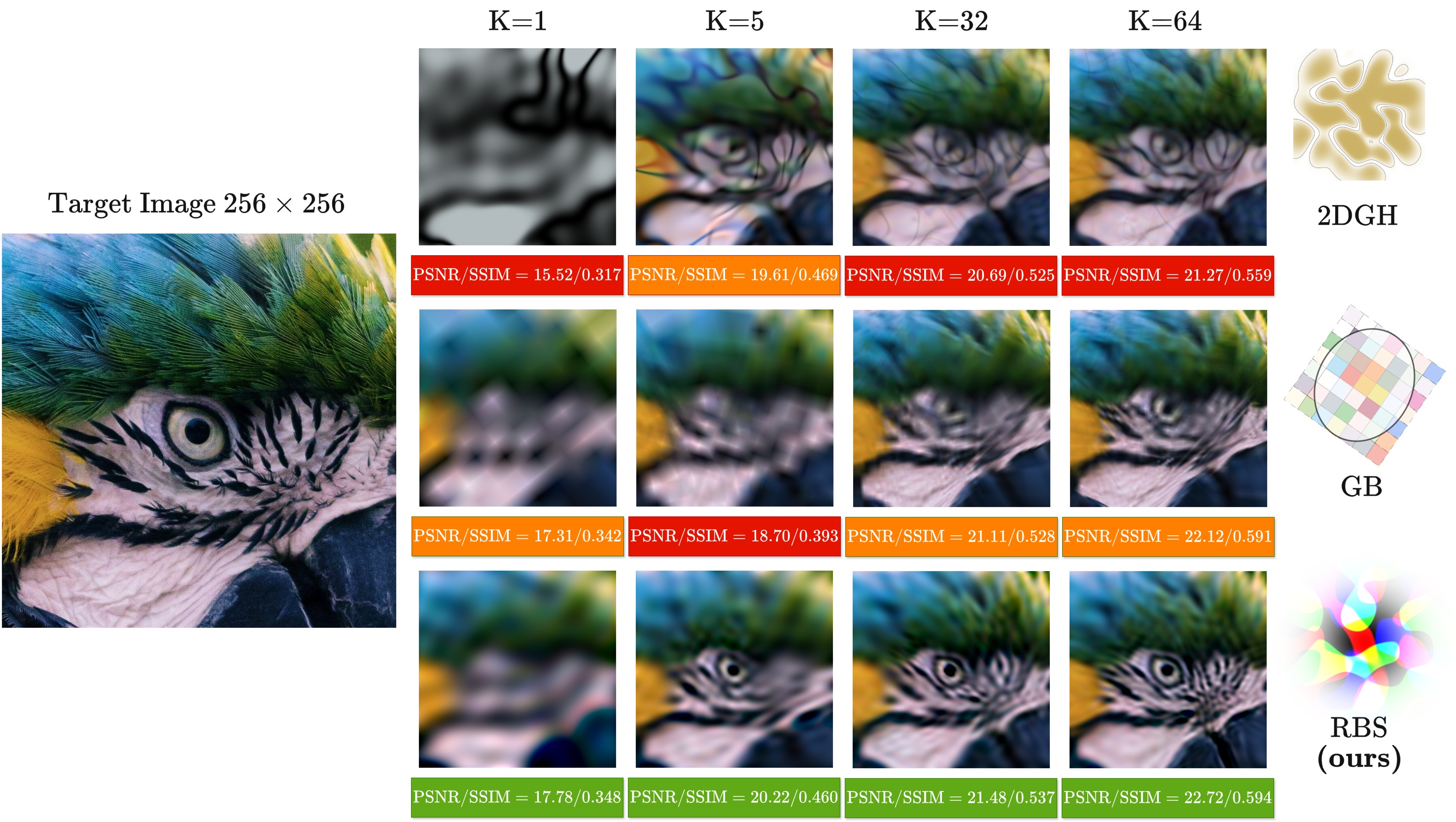}

\vspace{0.2cm} 
\newcommand{\bestcell}[1]{\cellcolor{green!25}\textbf{#1}}
\newcommand{\midcell}[1]{\cellcolor{orange!25}#1}
\newcommand{\worstcell}[1]{\cellcolor{red!25}#1}

\setlength{\tabcolsep}{3pt} 
\resizebox{\columnwidth}{!}{%
\begin{tabular}{l ccc ccc ccc}
\toprule
\multirow{2}{*}{Method} & \multicolumn{3}{c}{$\times 3$} & \multicolumn{3}{c}{$\times 6$} & \multicolumn{3}{c}{$\times 12$} \\
\cmidrule(lr){2-4} \cmidrule(lr){5-7} \cmidrule(lr){8-10}
& PSNR $\uparrow$ & SSIM $\uparrow$ & Time $\downarrow$ & PSNR $\uparrow$ & SSIM $\uparrow$ & Time $\downarrow$ & PSNR $\uparrow$ & SSIM $\uparrow$ & Time $\downarrow$ \\
\midrule
RBS  & \bestcell{29.98} & \bestcell{0.8836} & \bestcell{$\sim$307} & \bestcell{25.08} & \bestcell{0.7381} & \bestcell{$\sim$88} & \bestcell{21.71} & \bestcell{0.5893} & \bestcell{$\sim$45} \\
GB   & \midcell{29.68}  & \midcell{0.8779}  & \worstcell{$\sim$838} & \midcell{24.71}  & \midcell{0.7211}  & \worstcell{$\sim$282} & \midcell{21.48}  & \midcell{0.5679}  & \worstcell{$\sim$92} \\
2DGH & \worstcell{29.32}  & \worstcell{0.8748}  & \worstcell{$\sim$838} & \worstcell{24.51}  & \worstcell{0.7201}  & \worstcell{$\sim$282} & \worstcell{21.44}  & \worstcell{0.5654}  & \worstcell{$\sim$92} \\
\bottomrule
\end{tabular}%
}
\caption{\textbf{Alternative primitives.} \textbf{Top:} Reconstruction vs. primitive count ($K$). For equal capacity, RBS and 2DGH~\cite{2dgh} use $N=M=12$, matching a $13\times 13$ grid in Gaussian Billboards (GB)~\cite{gaussianbillboards}. \textbf{Bottom:} Results on Urban100~\cite{urban100} using EDSR\cite{Lim_2017_CVPR_Workshops}.}
\label{fig:rbs_vs_2dgh}
\end{figure}

\paragraph{Alternative primitives
.}
We adapted 2DGH~\cite{2dgh} and Gaussian Billboards (GB)~\cite{gaussianbillboards} to our rasterizer, retraining both from scratch (Fig.~\ref{fig:rbs_vs_2dgh}).
The single-primitive visualization ($K=1$, Fig.~\ref{fig:rbs_vs_2dgh} Top) highlights our divergence. 2DGH deforms its footprint but still emits a flat color; its arbitrary shape prevents exact boundary culling, while its flat color representation still forces the use of dense overlapping. As acknowledged~\cite{2dgh}, this makes 2DGH slower than standard GS. 
In contrast to 2DGH, GB replaces flat colors with a discrete learnable $N\times N$ grid, and Fig.~\ref{fig:rbs_vs_2dgh} shows how this approach outperforms 2DGH, \textit{already} proving that internal color complexity inherently beats complex shapes. However, GB (as well as GSTex~\cite{GStex}, which adapts grid resolution to primitive size) remains discrete. RBS instead models color \textit{continuously}, which perfectly suits the continuous nature of ASR, yielding superior accuracy (Fig.~\ref{fig:rbs_vs_2dgh}, Bottom).

\subsection{Ablation Study}
\label{sec:ablation}

\begin{figure}
\centering
\includegraphics[width=\columnwidth]{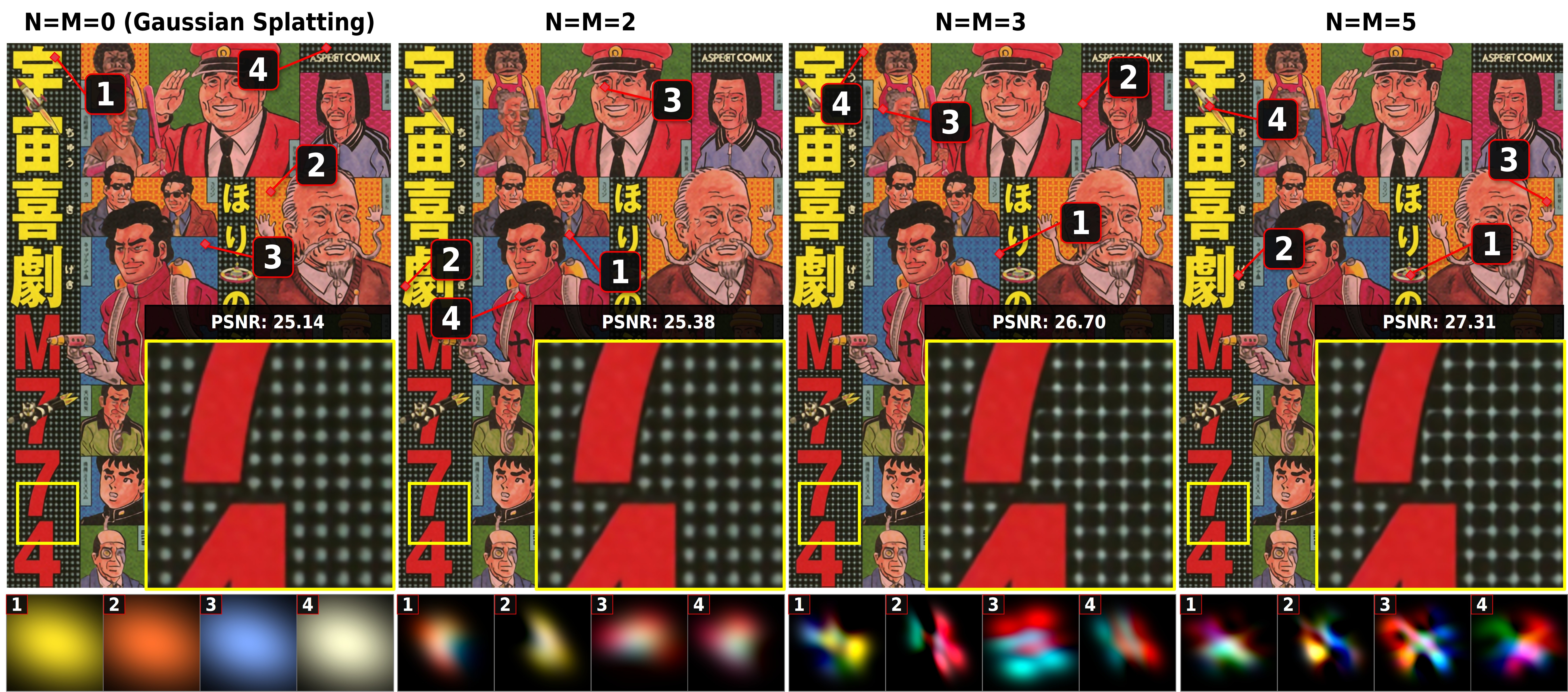}
\caption{\textbf{Impact of Brane degree.} Higher-order modes ($N=M \in \{2,3,5\}$) progressively capture sharper details and complex textures than standard Gaussians ($N=M=0$).}
\label{fig:comparison_models_with_different_max_polinomial_degree}
\end{figure}

\begin{table}[tb]
\centering

\resizebox{\columnwidth}{!}{
\begin{tabular}{ll cccc}
\toprule
\multirow{2}{*}{\textbf{Ablation Type}} & \multirow{2}{*}{\textbf{Metric}} & \multicolumn{4}{c}{\textbf{Scales}} \\
\cmidrule(lr){3-6}
& & \textbf{$\times 2$} & \textbf{$\times 4$} & \textbf{$\times 8$} & \textbf{$\times 12$} \\
\midrule

\rowcolor{gray!15}
\multicolumn{6}{l}{\textit{Feature Extractor: Advantage of using EDSR over RDN}} \\

\multirow{4}{*}{EDSR vs RDN} 
& Speed-up $\uparrow$ & 
\RankCellFour[+10.05\%]{10.05}{13.33} & \RankCellFour[+11.30\%]{11.30}{13.33} & \RankCellFour[+12.50\%]{12.50}{13.33} & \RankCellFour[\textbf{+13.33\%}]{13.33}{13.33} \\

& Mem. Saving $\uparrow$ & 
\RankCellFour[+0.58\%]{0.58}{13.84} & \RankCellFour[+2.21\%]{2.21}{13.84} & \RankCellFour[+6.90\%]{6.90}{13.84} & \RankCellFour[\textbf{+13.84\%}]{13.84}{13.84} \\

& PSNR $\Delta$ $\uparrow$ & 
\RankCellFour[-0.14\%]{-0.14}{0.14} & \RankCellFour[-0.10\%]{-0.10}{0.14} & \RankCellFour[\textbf{+0.00\%}]{0.00}{0.14} & \RankCellFour[-0.04\%]{-0.04}{0.14} \\

& SSIM $\Delta$ $\uparrow$ & 
\RankCellFour[-0.02\%]{-0.02}{0.13} & \RankCellFour[-0.09\%]{-0.09}{0.13} & \RankCellFour[-0.01\%]{-0.01}{0.13} & \RankCellFour[\textbf{+0.13\%}]{0.13}{0.13} \\

\midrule
\rowcolor{gray!15}
\multicolumn{6}{l}{\textit{Culling Strategy: Rasterization Speedup with $R_{\max}$ Enabled}} \\

\multirow{2}{*}{$R_{\max}$ Active} & Time Saving $\uparrow$ & 
\RankCellFour[+59.24\%]{59.24}{65.07} & \RankCellFour[\textbf{+65.07\%}]{65.07}{65.07} & \RankCellFour[+63.82\%]{63.82}{65.07} & \RankCellFour[+47.42\%]{47.42}{65.07} \\
& PSNR/SSIM $\Delta \uparrow$ & \RankCellFour[0.00\%]{0.00}{1.00} & \RankCellFour[0.00\%]{0.00}{1.00} & \RankCellFour[0.00\%]{0.00}{1.00} & \RankCellFour[0.00\%]{0.00}{1.00} \\

\midrule
\rowcolor{gray!15}
\multicolumn{6}{l}{\textit{Color Space: Performance impact of using YCbCr over RGB}} \\

\multirow{2}{*}{YCbCr vs RGB} 
& PSNR $\Delta$ $\uparrow$ & 
\RankCellFour[\textbf{-0.69\%}]{-0.69}{0.94} & \RankCellFour[-0.94\%]{-0.94}{0.94} & \RankCellFour[-0.81\%]{-0.81}{0.94} & \RankCellFour[-0.79\%]{-0.79}{0.94} \\

& SSIM $\Delta$ $\uparrow$ & 
\RankCellFour[\textbf{-0.07\%}]{-0.07}{1.00} & \RankCellFour[-0.59\%]{-0.59}{1.00} & \RankCellFour[-1.00\%]{-1.00}{1.00} & \RankCellFour[-0.92\%]{-0.92}{1.00} \\

\midrule
\rowcolor{gray!15}
\multicolumn{6}{l}{\textit{Hermite degree: Advantage of $N=M$ over Baseline ($N=M=0$)}} \\

\scriptsize $N=M=2$ 
& \scriptsize PSNR/SSIM $\Delta \uparrow$ & 
\scriptsize \RankInline[+0.05\%]{0.05}{0.64} / \RankInline[+0.01\%]{0.01}{0.44} & 
\scriptsize \RankInline[+0.06\%]{0.06}{0.64} / \RankInline[+0.07\%]{0.07}{0.44} & 
\scriptsize \RankInline[+0.00\%]{0.00}{0.64} / \RankInline[+0.00\%]{0.00}{0.44} & 
\scriptsize \RankInline[+0.04\%]{0.04}{0.64} / \RankInline[+0.07\%]{0.07}{0.44} \\
\cmidrule{1-6}

\scriptsize $N=M=3$ 
& \scriptsize PSNR/SSIM $\Delta \uparrow$ & 
\scriptsize \RankInline[+0.45\%]{0.45}{0.64} / \RankInline[+0.03\%]{0.03}{0.44} & 
\scriptsize \RankInline[+0.28\%]{0.28}{0.64} / \RankInline[+0.06\%]{0.06}{0.44} & 
\scriptsize \RankInline[+0.07\%]{0.07}{0.64} / \RankInline[+0.00\%]{0.00}{0.44} & 
\scriptsize \RankInline[+0.08\%]{0.08}{0.64} / \RankInline[+0.13\%]{0.13}{0.44} \\
\cmidrule{1-6}

\scriptsize $N=M=5$ 
& \scriptsize PSNR/SSIM $\Delta \uparrow$ & 
\scriptsize \RankInline[\textbf{+0.64\%}]{0.64}{0.64} / \RankInline[\textbf{+0.07\%}]{0.07}{0.44} & 
\scriptsize \RankInline[\textbf{+0.44\%}]{0.44}{0.64} / \RankInline[\textbf{+0.16\%}]{0.16}{0.44} & 
\scriptsize \RankInline[\textbf{+0.32\%}]{0.32}{0.64} / \RankInline[\textbf{+0.23\%}]{0.23}{0.44} & 
\scriptsize \RankInline[\textbf{+0.34\%}]{0.34}{0.64} / \RankInline[\textbf{+0.44\%}]{0.44}{0.44} \\
\cmidrule{1-6}

\scriptsize $N=M=7$ 
& \scriptsize PSNR/SSIM $\Delta \uparrow$ & 
\scriptsize \RankInline[+0.51\%]{0.51}{0.64} / \RankInline[+0.04\%]{0.04}{0.44} & 
\scriptsize \RankInline[+0.41\%]{0.41}{0.64} / \RankInline[+0.12\%]{0.12}{0.44} & 
\scriptsize \RankInline[\textbf{+0.32\%}]{0.32}{0.64} / \RankInline[+0.20\%]{0.20}{0.44} & 
\scriptsize \RankInline[\textbf{+0.34\%}]{0.34}{0.64} / \RankInline[+0.41\%]{0.41}{0.44} \\

\bottomrule
\end{tabular}
}
\caption{\textbf{Ablation study of the key components of RBS.}}
\label{tab:ablation_key_results_rbs}
\end{table}

\paragraph{Ablation on key components.}
Table~\ref{tab:ablation_key_results_rbs} ablates core designs on DIV2K~\cite{Timofte_2017_CVPR_Workshops} test set (full study in Supp.). Swapping RDN for EDSR trades minor quality for progressive efficiency gains (e.g., $+13.8\%$ memory at $\times 12$). Crucially, $R_{\max}$ culling accelerates rendering up to $65\%$ without quality loss, proving Branes resolve complexity without dense overlap. Shifting to YCbCr slightly degrades performance: our single MLP struggles to balance asymmetric distributions (vibrating Y vs. flat Cb/Cr) compared to correlated RGB. Increasing Hermite degrees to $N=M=5$ captures sharper details (Fig.~\ref{fig:comparison_models_with_different_max_polinomial_degree}), but predicting excessive modes ($N=M=7$) in a single pass hampers generalization and causes smooth-region artifacts, addressed below.

\begin{figure}[t]
\centering

\includegraphics[width=0.98\columnwidth]{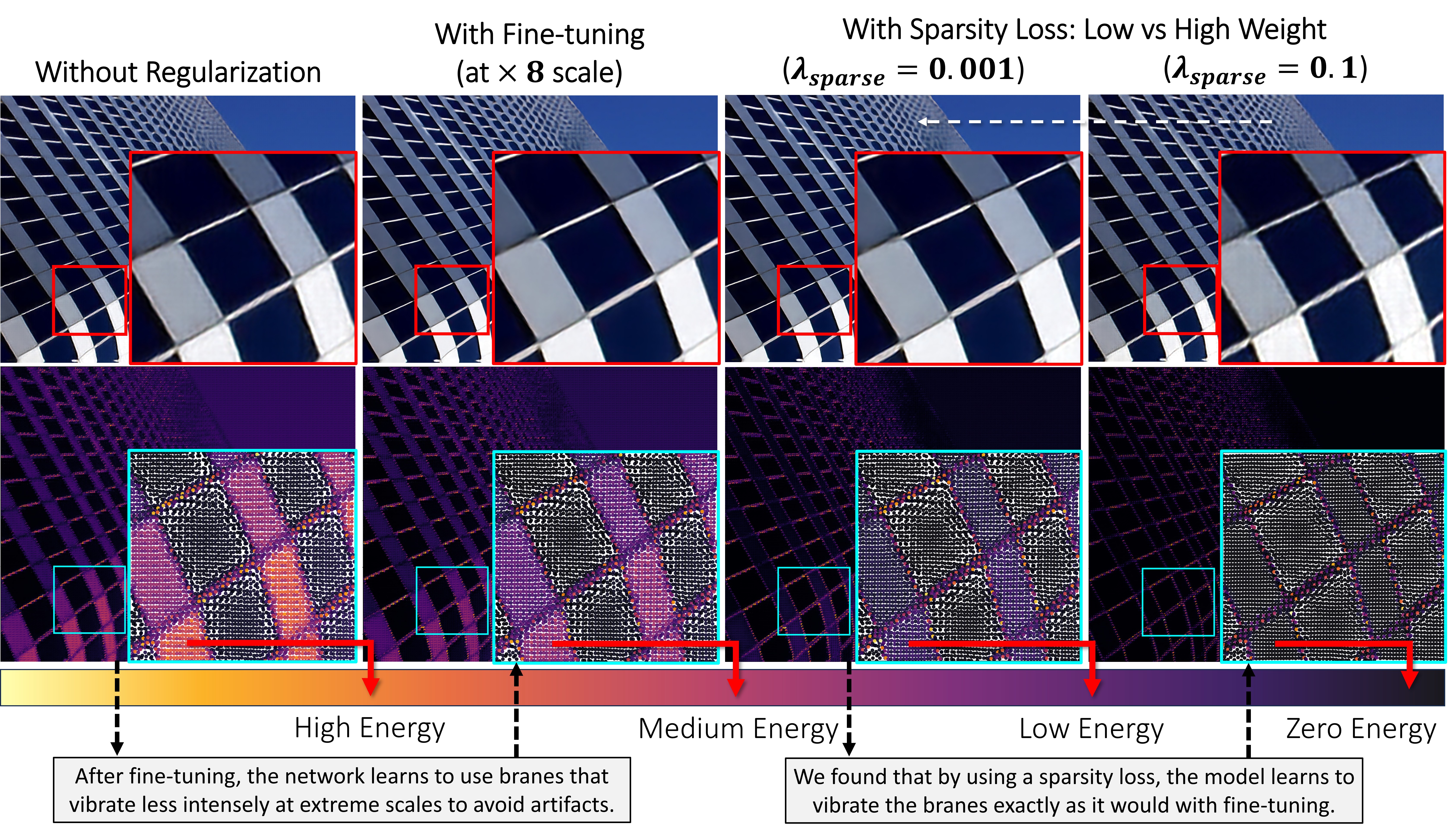}

\begin{table}[H]
\centering

\newcommand{\CC}[2][]{\if\relax\detokenize{#1}\relax\HL{#2}{0.0}{100.0}\else\HL[#1]{#2}{0.0}{100.0}\fi}

\setlength{\tabcolsep}{3.5pt}
\resizebox{\linewidth}{!}{%
\begin{tabular}{l cccccccccccc}
\toprule
\multirow{2}{*}{Model} & \multicolumn{12}{c}{Scale (PSNR / SSIM)} \\
\cmidrule(lr){2-13}
 & \multicolumn{2}{c}{$\times 2$} & \multicolumn{2}{c}{$\times 3$} & \multicolumn{2}{c}{$\times 4$} & \multicolumn{2}{c}{$\times 6$} & \multicolumn{2}{c}{$\times 8$} & \multicolumn{2}{c}{$\times 12$} \\
\midrule

w/o Regularization & \CC[\textbf{37.82}]{0.0} & \CC[0.9522]{20.0} & \CC[34.02]{24.0} & \CC[0.9007]{25.0} & \CC[32.00]{23.5} & \CC[0.8541]{22.7} & \CC[29.66]{29.4} & \CC[0.7844]{41.5} & \CC[28.23]{35.7} & \CC[0.7384]{54.5} & \CC[26.49]{41.7} & \CC[0.6847]{66.1} \\
w/ Fine-Tuning & \CC[\textbf{37.82}]{0.0} & \CC[0.9522]{20.0} & \CC[\textbf{34.08}]{0.0} & \CC[\textbf{0.9012}]{0.0} & \CC[\textbf{32.04}]{0.0} & \CC[\textbf{0.8546}]{0.0} & \CC[\textbf{29.71}]{0.0} & \CC[\textbf{0.7861}]{0.0} & \CC[\textbf{28.28}]{0.0} & \CC[\textbf{0.7414}]{0.0} & \CC[\textbf{26.54}]{0.0} & \CC[\textbf{0.6888}]{0.0} \\
Sparsity Loss ($\lambda_\text{sparse}=0.1$) & \CC[37.14]{100.0} & \CC[0.9514]{100.0} & \CC[33.83]{100.0} & \CC[0.8992]{100.0} & \CC[31.87]{100.0} & \CC[0.8524]{100.0} & \CC[29.54]{100.0} & \CC[0.7820]{100.0} & \CC[28.14]{100.0} & \CC[0.7359]{100.0} & \CC[26.42]{100.0} & \CC[0.6826]{100.0} \\
Sparsity Loss ($\lambda_\text{sparse}=0.001$) & \CC[37.54]{41.2} & \CC[0.9520]{40.0} & \CC[34.01]{28.0} & \CC[0.9003]{45.0} & \CC[31.99]{29.4} & \CC[0.8537]{40.9} & \CC[29.65]{35.3} & \CC[0.7842]{46.3} & \CC[28.23]{35.7} & \CC[0.7382]{58.2} & \CC[26.50]{33.3} & \CC[0.6852]{58.1} \\
Sparsity Loss ($\lambda_\text{sparse}=0.0005$) & \CC[37.79]{4.4} & \CC[\textbf{0.9524}]{0.0} & \CC[34.04]{16.0} & \CC[0.9008]{20.0} & \CC[32.02]{11.8} & \CC[0.8537]{40.9} & \CC[29.69]{11.8} & \CC[0.7860]{2.4} & \CC[28.27]{7.1} & \CC[0.7412]{3.6} & \CC[26.52]{16.7} & \CC[0.6883]{8.1} \\

\bottomrule
\\ 
\toprule

\multirow{2}{*}{Model} & \multicolumn{12}{c}{Time} \\
\cmidrule(lr){2-13}
 & \multicolumn{2}{c}{$\times 2$} & \multicolumn{2}{c}{$\times 3$} & \multicolumn{2}{c}{$\times 4$} & \multicolumn{2}{c}{$\times 6$} & \multicolumn{2}{c}{$\times 8$} & \multicolumn{2}{c}{$\times 12$} \\
\midrule

w/o Adaptive Mode Culling & \multicolumn{2}{c}{\CC[701]{100.0}} & \multicolumn{2}{c}{\CC[223]{100.0}} & \multicolumn{2}{c}{\CC[119]{100.0}} & \multicolumn{2}{c}{\CC[64]{100.0}} & \multicolumn{2}{c}{\CC[57]{100.0}} & \multicolumn{2}{c}{\CC[49]{100.0}} \\
w/ Adaptive Mode Culling & \multicolumn{2}{c}{\CC[\textbf{698}]{0.0}} & \multicolumn{2}{c}{\CC[\textbf{220}]{0.0}} & \multicolumn{2}{c}{\CC[\textbf{114}]{0.0}} & \multicolumn{2}{c}{\CC[\textbf{61}]{0.0}} & \multicolumn{2}{c}{\CC[\textbf{53}]{0.0}} & \multicolumn{2}{c}{\CC[\textbf{41}]{0.0}} \\

\bottomrule
\end{tabular}%
}
\end{table}

\caption{\textbf{Regularization and Mode Culling.} \textbf{Top:} Fine-tuning and sparsity loss reduce excessive vibration in smooth regions. \textbf{Middle:} Quantitative comparison of both methods. \textbf{Bottom:} Inference speedups from Adaptive Mode Culling.}
\label{fig:FT_VS_sparsityloss}
\end{figure}

\paragraph{Limitations and Regularization.}
As visualized (Fig.~\ref{fig:brane_energy_variation}), Branes vibrate intensely at larger scales to compensate for primitive sparsity. However, training strictly on $s \in \mathcal{U}(1, 4)$ fails to suppress these oscillations when generalizing to smooth regions at higher scales. Here, over-expressing modes injects spurious spatial frequencies, causing aliasing artifacts (Fig.~\ref{fig:FT_VS_sparsityloss} Top). Fine-tuning on $s \in \mathcal{U}(1, 8)$ provides an \textit{implicit} solution: exposed to smoother targets, the model naturally dampens coefficients ($c_{n,m} \to 0$ for $n+m > 0$) in flat areas, recovering performance (Fig.~\ref{fig:FT_VS_sparsityloss} Middle).
To 
prove this energy reallocation, we tested an \textit{explicit} route by adding a sparsity loss to the $\mathcal{L}_1$ objective: $\mathcal{L}_\text{sparse} = \lambda_\text{sparse} \frac{1}{K} \sum_{k=1}^{K} \sum_{n+m>0} \| c_{n,m,k} \|_1$. While an aggressive penalty ($\lambda_\text{sparse}=0.1$) stifles details, a tuned weight ($\lambda_\text{sparse}=0.0005$) yields results virtually identical to the fine-tuned model. This supports that fine-tuning 
induces the exact ``energy awareness'' that $\mathcal{L}_\text{sparse}$ enforces directly.


\paragraph{Adaptive Mode Culling.}
Fine-tuning and $\mathcal{L}_\text{sparse}$ train the model to zero unneeded coefficients, which our CUDA rasterizer can exploit via \textit{Adaptive Mode Culling}. If a mode's RGB coefficients are $<10^{-4}$, it skips Hermite evaluation and color accumulation. By filtering this near-zero noise, it primarily yields modest speedups (given our already highly efficient rasterizer) with zero quality loss (Fig.~\ref{fig:FT_VS_sparsityloss}, bottom).

\section{Conclusion}

We introduced Resonant Brane Splatting for Arbitrary-Scale Super-Resolution, proposing the Brane primitive. By internally modeling complex colors, Branes avoid the dense overlap required by Gaussian Splatting. This saves memory and accelerates inference, with rasterization further optimized via the quantum turning point. Though sparsity and fine-tuning mitigate extreme-scale smooth-region oscillations, perfect generalization remains future work. Overall, RBS balances expressive reconstruction with efficiency.
\clearpage

{
    \small
    \bibliographystyle{ieeenat_fullname}
    \bibliography{main}
}

\end{document}